\documentclass[preprint,12pt]{elsarticle}
\usepackage[utf8]{inputenc}
\usepackage[T1]{fontenc}
\usepackage[english]{babel}
\usepackage{indentfirst}
\usepackage{enumitem}
\usepackage{amssymb}
\usepackage{amsmath}
\usepackage{amsfonts}
\usepackage{multicol}
\usepackage{mathrsfs}
\usepackage{stmaryrd}
\usepackage{lscape}
\usepackage{colortbl}
\usepackage{fancyhdr}
\usepackage{tabularx,environ}
\usepackage{appendix}
\usepackage{array}
\usepackage{multirow}
\usepackage{colortbl}
\usepackage{epsfig}
\usepackage{indentfirst}
\usepackage{amsmath}
\usepackage{amsfonts}
\usepackage{amssymb}
\usepackage{calrsfs}
\usepackage{caption}
\usepackage{subcaption}
\usepackage{color}
\usepackage{cancel}
\usepackage[table,dvipsnames]{xcolor}
\usepackage[a4paper,left=2.7cm,right=2.7cm,top=2.7cm,bottom=2.7cm]{geometry}
\usepackage{url}
\usepackage[hidelinks]{hyperref}
\usepackage{soul}
\newcolumntype{C}[1]{>{\centering\arraybackslash}m{#1}}
\newcolumntype{R}[1]{>{\raggedleft\arraybackslash }b{#1}}
\newcolumntype{L}[1]{>{\raggedright\arraybackslash }b{#1}}

\setstcolor{red}

\usepackage{amssymb}

\journal{Arxiv}

\begin{document}

\begin{frontmatter}



\title{Hidden-Variables Genetic Algorithm for Variable-Size Design Space Optimal Layout Problems with Application to Aerospace Vehicles}


\author[inst1,inst2]{Juliette Gamot}

\affiliation[inst1]{organization={ONERA/DTIS},
            addressline={Université Paris-Saclay}, 
            city={Palaiseau},
            country={France}}

\author[inst1]{Mathieu Balesdent}
\author[inst1]{Arnault Tremolet}
\author[inst1]{Romain Wuilbercq}
\author[inst2]{Nouredine Melab}
\author[inst2]{El-Ghazali Talbi}

\affiliation[inst2]{organization={INRIA},
            addressline={Université de Lille}, 
            city={Villeneuve d'Ascq},
            country={France}}

\begin{abstract}
The optimal layout of a complex system such as aerospace vehicles consists in placing a given number of components in a container in order to minimize one or several objectives under some geometrical or functional constraints. 
This paper presents an extended formulation of this problem as a variable-size design space (VSDS) problem to take into account a large number of architectural choices and components allocation during the design process. As a representative example of such systems, considering the layout of a satellite module, the VSDS aspect translates the fact that the optimizer has to choose between several subdivisions of the components. For instance, one large tank of fuel might be placed as well as two smaller tanks or three even smaller tanks for the same amount of fuel. 
In order to tackle this NP-hard problem, a genetic algorithm enhanced by an adapted hidden-variables mechanism is proposed. This latter is illustrated on a toy case and an aerospace application case representative to real world complexity to illustrate the performance of the proposed algorithms. The results obtained using the proposed mechanism are reported and analyzed.

\end{abstract}




\end{frontmatter}


\section{Introduction}
\label{sec:intro}
\subsection{General context}
The field of Multidisciplinary Design and Optimization introduces a wide range of optimization techniques capable to generate innovative preliminary design solutions for various complex systems including automotive, satellites, reusable launchers, etc. \cite{deremaux} The design process of aerospace concepts requires a strong coupling between all the disciplines involved in their preliminary sizing. While a number of engineering fields such as aerodynamics, propulsion, structure and flight mechanics are often tightly integrated at the early design stage, the design of the internal layout of future systems \cite{deremaux} is, however, often put aside and is rarely part of a fully integrated design process. In fact, the internal layout problem is often solved by hand or by a set of simple heuristic rules (\textit{e.g.} expert system) able to mimic the cognitive process of experts. For instance, in the process of designing a new satellite platform, engineers must take into account subsystems' incompatibility and then position each component to guarantee a number of mission requirements. However, by the fact that this process is performed in the last steps of the design procedure, it does not guarantee that an overall optimal solution has been identified at preliminary design stages. Indeed, the arrangement of components often has a first-order impact on the performance of the system (\textit{e.g.} for flying qualities of an aircraft) and can thus be critical to the feasibility of a concept. The main concern is thus the automated optimal design of the internal layout of complex systems that makes it compatible with the inclusion in the early design phases.

The optimal layout design of a system consists in placing a set of components (various instruments and equipment) into an available space. The layout optimization often involves minimizing or maximizing one or several objective(s) (\textit{e.g.} minimize the inertia, the volume, the location of the center of gravity) while satisfying multiple constraints. For illustration purposes, in this paper, the layout optimization of satellite module is derived, without loss of generality. This problem is a representative example of layout optimization problems \cite{liu,cui2019,wang2009} and its formulation will be extended in this paper. 
Indeed, in most of real-world engineering design problems, many architecture choices have to be made (\textit{e.g.} number of engines for an aircraft, number of tanks of fuel for a satellite, etc.) and are often represented as discrete variables in addition to other continuous design variables. Furthermore, the choices made in terms of architecture or topology have a critical impact on the problem definition (number of variables to be optimized, number of constraints). For instance, a possible choice would be either to place in the container two small tanks of fuel or a large one. The number of components is consequently not the same in both cases and hence, the number of decision variables (\textit{e.g.} the positions of the tanks of fuel for instance) will vary from one sub-problem to another. Allowing the number of components to vary during the optimization may provide a better solution at the end. 
Consequently, the purpose is to provide a catalog of all the components which can possibly be chosen to be part of the system layout and let an optimization algorithm find the best combinations of components, as well as their positions in the container.
This type of problems is referred to in the literature as both mixed-variable problems and variable-size design space (VSDS) problems \cite{pelamatti}. Thus, in this paper, a new formulation of the optimal layout problem is proposed including the VSDS aspect which is, to the best of our knowledge, not covered in the literature. Then, a method dedicated to this kind of problems for optimal layout is proposed and associated to a genetic algorithm (GA). The algorithms proposed to solve such problems are described in Section \ref{sec:stateV}.

\begin{figure}[ht]
    \centering
    \includegraphics[width = 14cm]{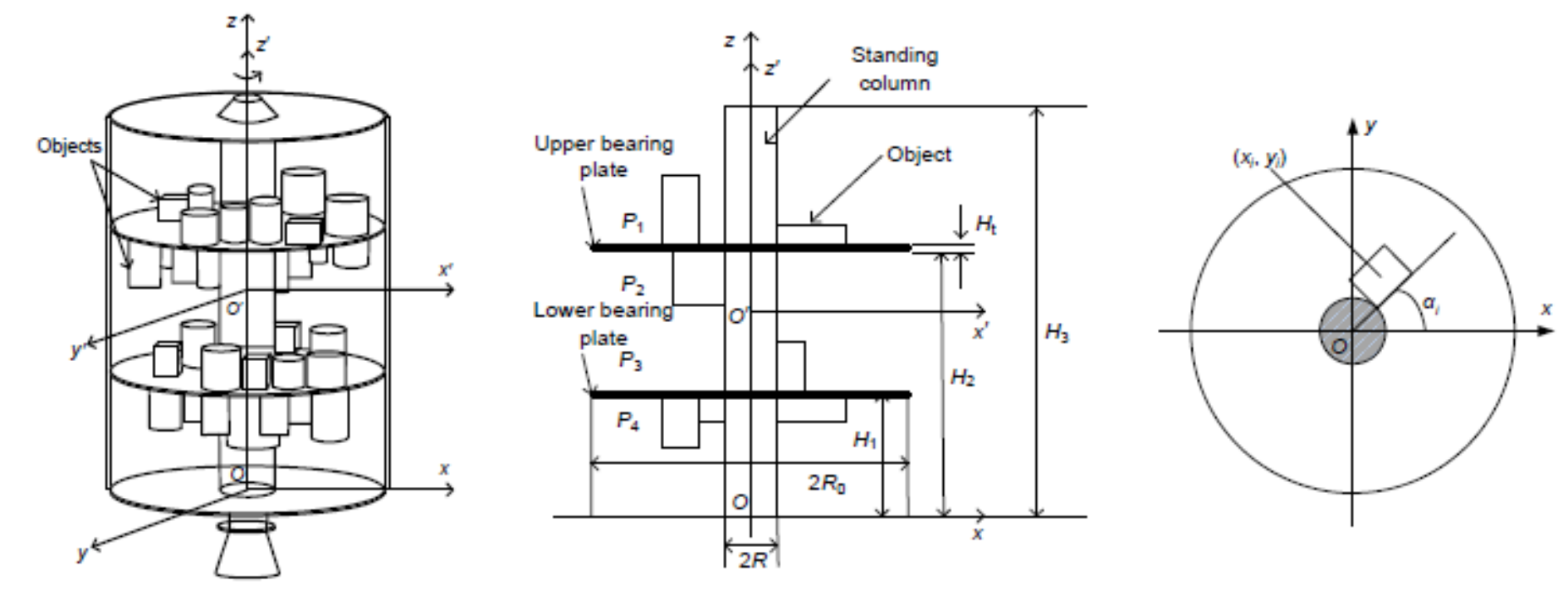}
    \caption{Different views of the simplified representation of the satellite module \cite{liu}.}
    \label{fig:liusat}
\end{figure}

In order to solve this problem, the simplified model of the international commercial communication satellite module (INTELSAT-III) is considered as a benchmark \cite{liu}. Figure \ref{fig:liusat} shows different views of this satellite module. 

\subsection{Optimal layout problem}

To illustrate the proposed mechanism, a satellite problem is presented. Indeed, the problem that is most of the time tackled in the literature is a fixed-size design space problem and consists in finding the layout that minimizes the inertia of the whole system composed with $N$ cuboid and cylindrical components on two bearing plates (or four surfaces) while respecting certain geometric constraints (\textit{e.g.} no overlapping between components, accurate placement of the center of gravity of the module). According to the studies \cite{liu,cui2019}, $N$ varies from a few dozens to one hundred. 

Generally, the optimal layout problem can be expressed as follows:
\begin{equation}
\left\{
\begin{array}{llll}
\text{Minimize } f(\textbf{x},\textbf{z}) \\
\text{where: } \textbf{x} \in \mathbb{R}^{n_{x}}, \textbf{z} \in \mathbb{Z}^{n_{z}}  \\
\text{subject to :}\\
\textbf{h(x,z)} = 0 \\
\textbf{g(x,z)} \leq 0
\end{array}
\right.
\label{eq:eq1}
\end{equation}

where $f$ is the objective function taking continuous design variables $\textbf{x}$ and discrete design variables $\textbf{z}$ as arguments. $\textbf{h}$ and $\textbf{g}$ stand for respectively equality and inequality constraints.

Regarding the satellite module layout problem, the continuous design variables $\textbf{x}$ are most of the time the positions of the components on a given surface and the discrete design variables $\textbf{z}$ are the orientations of the several components (\textit{e.g} cuboid, often 0 or 90 degrees) or the number of the plate on which the components have to be placed. While minimizing the inertia of the system, the following constraints must be satisfied: no overlapping between components and between components and the container (equality constraints), and the center of gravity of the module must be positioned accurately given a tolerance (inequality constraints).

In the literature, the number of components is set at the beginning of the study and a plate number is assigned to each of the components \cite{liu,cui2019}. This problem belongs to the category of "Fixed-size variable design space problems" \cite{pelamatti}. The algorithms that can be used to solve such problems are described in Section \ref{sec:stateF}. 

In this paper, the previous formulation is extended in order to let the optimizer find the best combination of components, instead of dealing with a fixed set of components that have to be place in the container. This new formulation of the optimal layout problem is detailed in Section \ref{sec:problem}. The optimizer has now to optimize the list of components to place in the container as well as their positions and orientations. The extended problem belongs to the VSDS class of problems. The methods that can be used to solve it are reviewed in Section \ref{sec:stateV}.

The main contributions of the paper are as follows. The formulation of the problem \ref{eq:eq1} is extended in order to take into account the VSDS aspect. To solve it, a mechanism entitled "Hidden-Variables Mechanism" is adapted from \cite{abdelkhalik2011}. The resulting adapted algorithms are applied to the extended satellite module optimal layout problem as a representative example of optimal layout problems. Moreover, the constraints handled in the problem described above are essentially geometrical (\textit{e.g.} center of mass, overlap). Functional constraints must be added in order to describe the problem more realistically if real-world problems are meant to be tackled. For instance, some components are incompatible for functional reasons and must be placed at a certain distance apart. The functional constraints can describe any requirements that the system needs to behave properly (\textit{e.g.} existence of a radiation threshold in the container).

Finally, a performance study on the compactness of the layout container (corresponding to the occupation rate of the container by the components) is conducted.  Most of the time in the existing literature, the compactness is set during the whole study, and often to a low value. The more the compactness increases, the more the constraints might be difficult to handle.

The rest of the paper is organized as follows: Section \ref{sec:state} reviews the existing methods either for fixed-size and variable-size design space problems.  Section \ref{sec:problem} introduces the new mathematical formulation of the variable-size layout optimization problem. In Section \ref{sec:algo}, the proposed mechanism is described as well as its application to a GA. Section \ref{sec:toy} illustrates the proposed algorithms on a toy case. Finally, the proposed methods are applied in Section \ref{sec:sat} to a real-world benchmark test case and the results are analyzed.

\section{State of the Art}
\label{sec:state}
\subsection{Review of the fixed-size design space methods}
\label{sec:stateF}
The methods developed in the literature in order to tackle the problem formulated in the system of equations \ref{eq:eq1} are reviewed in this section.
This type of problems belongs to the class of NP-hard problems due to its mathematical complexity, and is therefore not solvable by exact optimization methods \cite{wang2009}. 
During the past 20 years, various methods have been developed in order to solve layout optimization problems first defined by Cagan \textit{et al.} \cite{cagan}. In \cite{singh}, a number of approaches used to solve the facility layout problems (mostly meta-heuristics) are reviewed. Meller \textit{et al.} introduced a simulated annealing algorithm to solve facility layout problems \cite{meller}. In \cite{jacquenot}, Jacquenot used a genetic algorithm (GA) to address multi-objective layout problems adding a separation algorithm in order to facilitate the treatment of overlapping constraints. Liu \textit{et al.} proposed a Multi-Objective Particle Swarm Optimization algorithm to solve unequal-area facility layout problems \cite{liu2018}. In \cite{derakhshan}, a Covariance Matrix Adptation - Evolution Strategy (CMA-ES) is compared to a Particle Swarm Optimization (PSO) algorithm and a GA to solve layout optimization problem, giving advantage to the CMA-ES algorithm. Hasan \textit{et al.} \cite{hasan} proposed an extended Ants Colony Optimization (ACO) algorithm to solve Dynamic Layout problems.

\indent With the increase in the number of design objectives, constraints or design variables, other techniques have been proposed in order to improve the computational performance of evolutionary algorithms (EAs). Among them, cooperative co-evolutionary algorithms first introduced by Potter and De Jong \cite{potter94} and based on the \textit{divide and conquer} method along with the biological model of co-evolution of cooperating species have been developed. Ma \textit{et al.} recently proposed a survey on cooperative co-evolutionary algorithms \cite{ma2018}. Those algorithms were successfully applied to satellite layout optimization \cite{cui2019,wang2009,hong2010}. \\
\indent Metaheuristics methods have also been often hybridized between themselves \cite{lim2016} where a GA is hybridized with differential evolution, artificial bee colony and PSO in order to optimize layouts for multi-robot cellular manufacturing systems. EAs can also be hybridized with other methods as in \cite{li2016} where heuristic rules are coupled with EAs like PSO and ACO and applied to the layout optimization of a satellite module. Liu \textit{et al.}  \cite{liu2012} tried a Multi-agent GA for Constrained Layout Optimization in Satellite Cabin problems defining evolutionary operators adapted to a multi-agent environment. 

\indent Metaheuristics have also been hybridized with heuristic techniques especially for the balance circular packing problem which consists in placing equal or non-equal radius circular components in a circular container while minimizing the congestion along with placing the center of mass of the system of circles. In \cite{xu}, a GA optimizes the order of the components to place while a greedy algorithm based on the order-based positioning technique and heuristic rules construct the layout component by component. The same authors then extended the placement heuristic rules to parallelepipedic items in \cite{xu2010}. In \cite{oliveira}, EAs are also hybridized with a new load balancing heuristic to tackle the optimal layout satellite problem.

\indent Other methods than metaheuristics have been applied to layout optimization problems even if they are less investigated.
Tarkesh \textit{et al.} presented an approach based on multi-agent system to solve a facility layout problem where the procedure can be assimilated to a Markov chain and modeled as a Markov Decision Process \cite{tarkesh2009}. Burggräf \textit{et al.} conducted a study on the use of machine learning as resolution techniques for facility layout problems in \cite{burggraf2021}. Unsupervised learning, supervised learning and reinforcement learning (RL) are explored to solve these problems. More specifically, Tsuchiya \textit{et al.} \cite{tsuchiya1996} focused on a neural network approach to optimize the layout of $N$ components on $N^2$ locations. 
Despite the small amount of articles using RL to tackle aerospace vehicles optimal layout problems, some recent articles use this method to address chip floorplanning which is a similar layout problem. 
In \cite{vashisht}, a cyclic combination of RL and simulated annealing (SA) is proposed to solve integrated circuit placement. This cyclic application of RL and SA is used to provide a better initialization for SA. 
In \cite{mirhoseini}, the authors developed a deep RL approach capable to learn from its experience of achieving the layout of the chip in order to become better and faster. A trained agent places the macros one by one on the chip and by the end obtains a reward which is function of the wirelength, the congestion and the density of the chip. Moreover, the authors created a neural network architecture in order to predict reward on new lists of components and use it as the encoder layer of their policy. 
In \cite{cheng}, a joint learning method for solving both placement and routing problems is detailed. It integrates reinforcement learning with a gradient-based optimization scheme. As in \cite{mirhoseini}, a policy network learns how to maximize the rewards.

\subsection{Review of the variable-size design space methods}

\label{sec:stateV}
As the problem formulated in the system of equations \ref{eq:eq1} will be extended to become a VSDS problem, the methods designed to handle such problems are reviewed in this section.

Most  of  the  aforementioned  algorithms are usually designed to handle fixed-size design space problems even though some of them can address the mixed-variable problems. Some previous research has been conducted on VSDS problems.
Pelamatti \cite{pelamatti} extended Bayesian Optimization (BO) algorithms to mixed-variable and VSDS problems. However, BO algorithms are used for small numbers of design variables (typically lower or equal to 10) and are not applicable for layout optimization problem in which a few hundreds optimization variables have to be handled \cite{pelamatti}. Dasgupta \textit{et al.} \cite{dasgupta} developed a Structured GA which utilizes a multi-layer structure for the chromosome initially designed to tackle non-stationary function optimization. Gentile \textit{et al.} \cite{gentile} used the same idea to tackle VSDS problems applied to satellite tracking problems.
Hutt \textit{et al.} \cite{hutt2007} created a new crossover operator used with a GA in order to  perform meaningful crossover between variable-length chromosome. 
In \cite{abdelpop}, a Dynamic-Size Multiple Population Genetic Algorithm (DSMPGA) is described. A certain number of initial subpopulations of fixed-size design spaces are generated and classical genetic operators are carried out at each generation. Finally, the process leads to an increase in the size of subpopulations of more fit solutions and a decrease in the size of subpopulations of less fit solutions. 
Recently, a self-adaptive PSO using the self-adaptive mechanism strategy as well as a parameter self-adaptative mechanism has been proposed for solving large-scale feature selection problems \cite{xue,xue2}.

Another idea to solve VSDS problems is the Hidden-Genes GA (HGGA) first introduced by Abdelkhalik in \cite{abdelkhalik2011,ellithy}. In HGGA, all chromosomes have the same length but only some of their genes are expressed during the objective and constraints evaluations. \\
To the best of our knowledge, none of the previous methods have been applied to optimal layout problems. Moreover, BO as well as DSMPGA are not adapted to large combinatory VSDS problems \cite{pelamatti}. The Structured GA requires some engineering expertise on the problem tackled regarding its encoding and is consequently not adapted to address various VSDS layout problems.
The proposed mechanism in this paper is inspired and adapted from HGGA. It is the most versatile method and requires very few engineering expertise on the problem tackled regarding its encoding, unlike the Structured Genetic Algorithm. It also allows to tackle large-sized problems including several thousands of subproblems unlike the Bayesian Optimization and the DSMPGA.

\section{Description of variable-size optimal layout problem}
\label{sec:problem}

\subsection{Illustrative example}
\begin{figure}[ht]
    \centering
    \includegraphics[width=10cm]{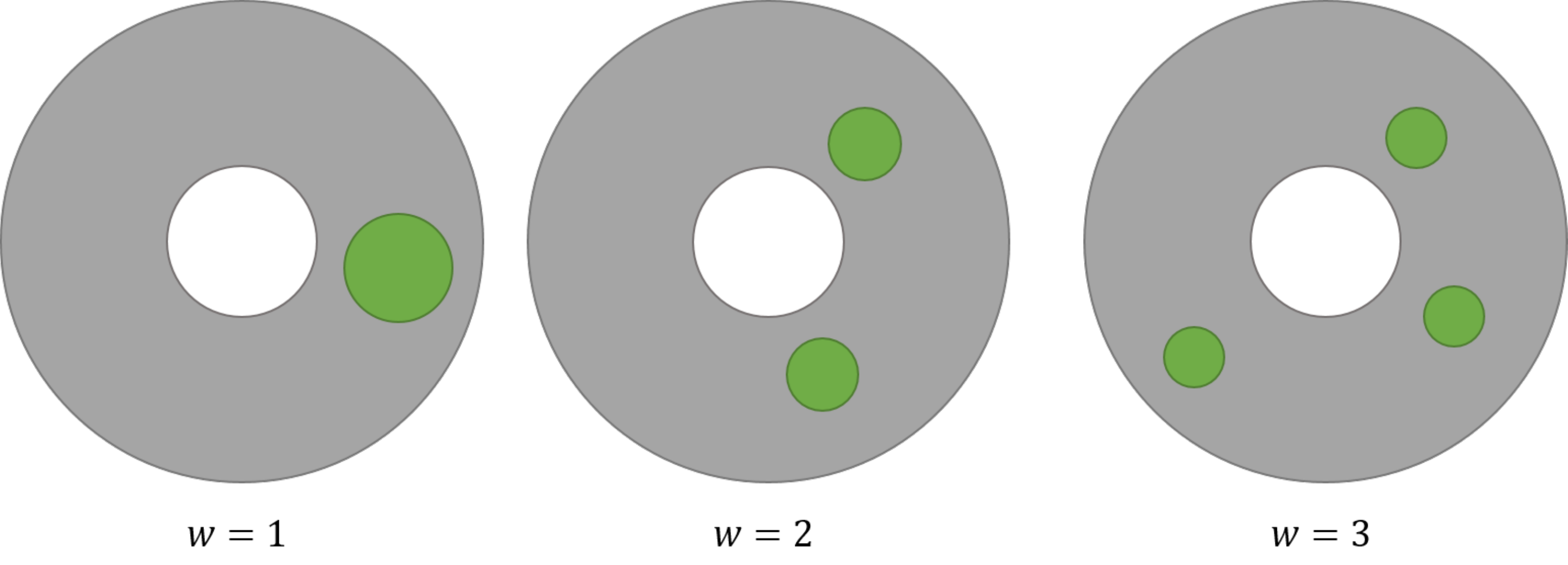}
    \caption{Three possible layouts for the same component to be placed with $w$ the number of tanks.}
    \label{fig:configex}
\end{figure}

The classical optimal layout problems described in the literature have to be extended as mentioned in Section \ref{sec:intro} to address general layout design and subdivision optimization. The main extension consists in dealing with VSDS features. As an example, a very simple case is considered in which only a tank of fuel has to be placed in the container. Several architecture choices might therefore be possible, without loss of generality. For instance, for a given amount of fuel, one large tank may be placed, as well as two smaller tanks, or even three smaller tanks of fuel. Figure \ref{fig:configex} shows the three possible layouts for the same problem definition corresponding to this example. 
The possible decompositions of the components are introduced by the designer and the optimizer will draw from it. Thus, the subdivisions of the components are optimized along with the layout. 

In this paper, in order to propose a general mathematical formulation of the mixed-variable VSDS problem, the design variables, the objective function and the constraints considered are described.

\subsection{Design variables}
The design variables can be split into 3 categories \cite{pelamatti}:
\begin{itemize}
    \item Continuous variables: \textbf{x}. These variables refer to real numbers, each of them defined in a given interval. Example: position and orientation of the components.
    \item Discrete variables: \textbf{z}. These variables are non-relaxable variables, defined among a finite set of possibilities. Example: the number of the bearing plate on which a component will be positioned.
    \item Dimensional variables: \textbf{w}. These variables are also  non-relaxable variables defined among a finite set of possibilities. According to their values, the number and type of continuous and discrete variables that have to be optimized in the problem may vary and also the considered constraints. Example: the subdivision choices.
\end{itemize}
Compared to the fixed-size design space problem definition \cite{liu,cui2019,liu2012}, the dimensional variables are added in order to translate the VSDS features to the optimal layout problem. The dimensional variables are illustrated in Figure \ref{fig:configex}: to each value of $w$ corresponds a subdivision of the components. If $N$ components must be placed and each of them has $n_i$ subdivisions (for $i \in \{1,...,N\}$), then the number of configurations corresponds to the cardinality of the set of achievable values of $w$, $F_w$, and is defined by:
\begin{equation*}
    Card(F_w)=\prod_{i=1}^N n_i
\end{equation*}

\subsection{Objective function}
\label{sec:inertia}
One of the most popular objective function to minimize in optimal layout problems is the inertia of the whole module calculated at the current center of mass \cite{liu2008}. The three coordinates systems initially proposed by Sun \textit{et al.} \cite{sun} are adopted in order to calculate the total inertia of the module:
\begin{itemize}
    \item $O''x''y''z''$: the coordinate system related to each component. $O''$ corresponds to the center of inertia of the component and the axes are defined as shown in Figure \ref{fig:coordinates} as the symmetry axes of the components.
    \item $O'x'y'$: the coordinates system related to the system of components. $O'$ represents the current centroid of the system of components and the axes are defined as the symmetry axes of the module. 
    \item $Oxy$: the coordinates system related to the module. $O$ is the geometric center of the container and the axes are its symmetry axes.
Figure \ref{fig:coordinates} illustrates the three systems of coordinates \cite{liu}.
\begin{figure}[ht]
    \centering
    \includegraphics[width=10cm]{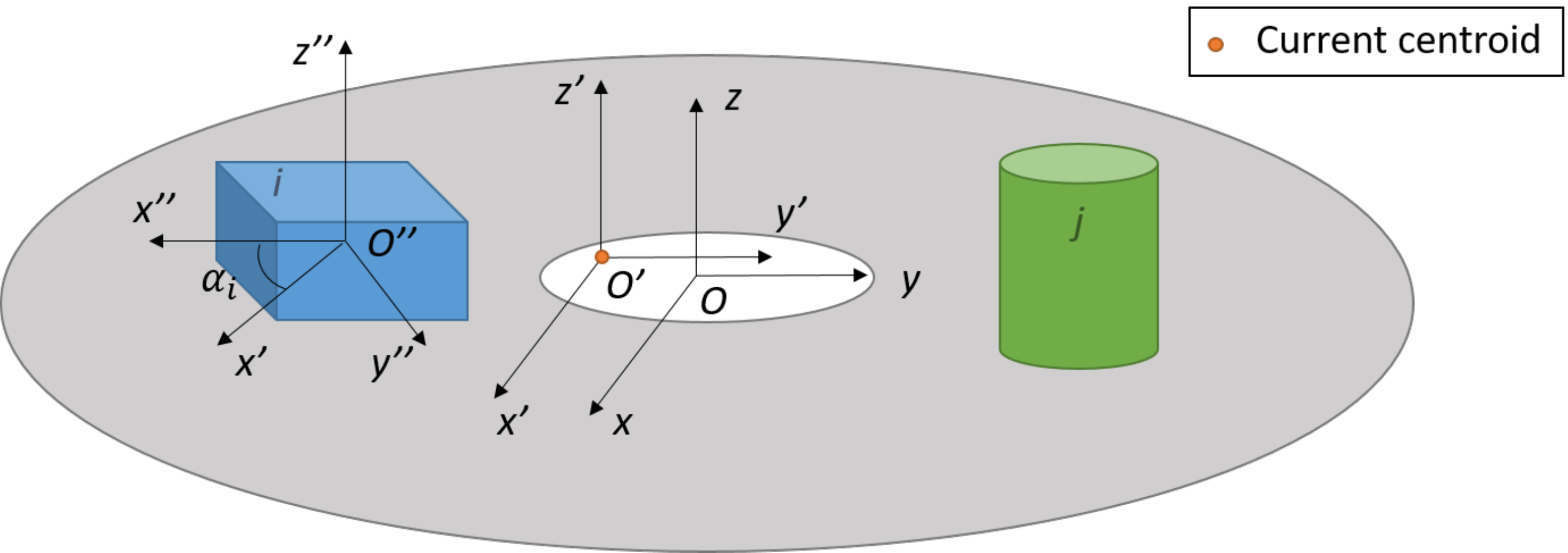}
    \caption{Sketch of the coordinates systems, \cite{liu}}
    \label{fig:coordinates}
\end{figure}

\end{itemize}
The total inertia is calculated in the $O'x'y'z'$ system of coordinates and is defined as:
\begin{equation*}
    f(\textbf{x},\textbf{z},\textbf{w})=I_{x'}+I_{y'}+I_{z'}
\end{equation*}

Each component has a solid inertia which depends on its shape and is expressed in its $O''x''y''z''$ coordinates system. The inertias along the three axes are noted $I_{x''i}$, $I_{y''i}$ and $I_{z''i}$. Thus, the inertia of the component in the system of coordinates $Oxyz$ is: 
\begin{equation*}
    I_{xi} = I_{x''i}\cos{\alpha_i}^2+I_{y''i}\sin{\alpha_i}^2+m_i(c_{i_y}^2+c_{i_z}^2)
\end{equation*}
\begin{equation*}
    I_{yi} = I_{y''i}\cos{\alpha_i}^2+I_{x''i}\sin{\alpha_i}^2+m_i(c_{i_x}^2+c_{i_z}^2)
\end{equation*}
\begin{equation*}
    I_{zi} = I_{z''i}+m_i(c_{i_x}^2+c_{i_y}^2)
\end{equation*}
where $m_i$ is the mass of the component \textit{i}, $c_{i_x}, c_{i_y}, c_{i_z}$ are the coordinates of its center of inertia in the system of coordinates $Oxyz$ and $\alpha_i$ its orientation.

A configuration $k$ where $k \in \{1,...,Card(F_w)\}$ is now considered, composed of $N_k$ components.
The Huygens theorem is used in order to express the inertia of this whole system of components in the system of coordinates $O'x'y'$. The coordinates of the current centroid of the system of components are noted $(x_c,y_c,z_c)$ and $m=\sum_{i=1}^{N_k}m_i$.
\begin{equation*}
    I_{x'}=\sum_{i=1}^{N_k} I_{xi} - m(y_c^2+z_c^2)
\end{equation*}
\begin{equation*}
    I_{y'}=\sum_{i=1}^{N_k} I_{yi} - m(x_c^2+z_c^2)
\end{equation*}
\begin{equation*}
    I_{z'}=\sum_{i=1}^{N_k} I_{zi} - m(x_c^2+y_c^2)
\end{equation*}

$I_{x'}$, $I_{y'}$ and $I_{z'}$ are the inertia of the module along each axis and calculated at the point $O'$.

\subsection{Constraints}
\begin{figure}[ht]
    \centering
    \includegraphics[width=10cm]{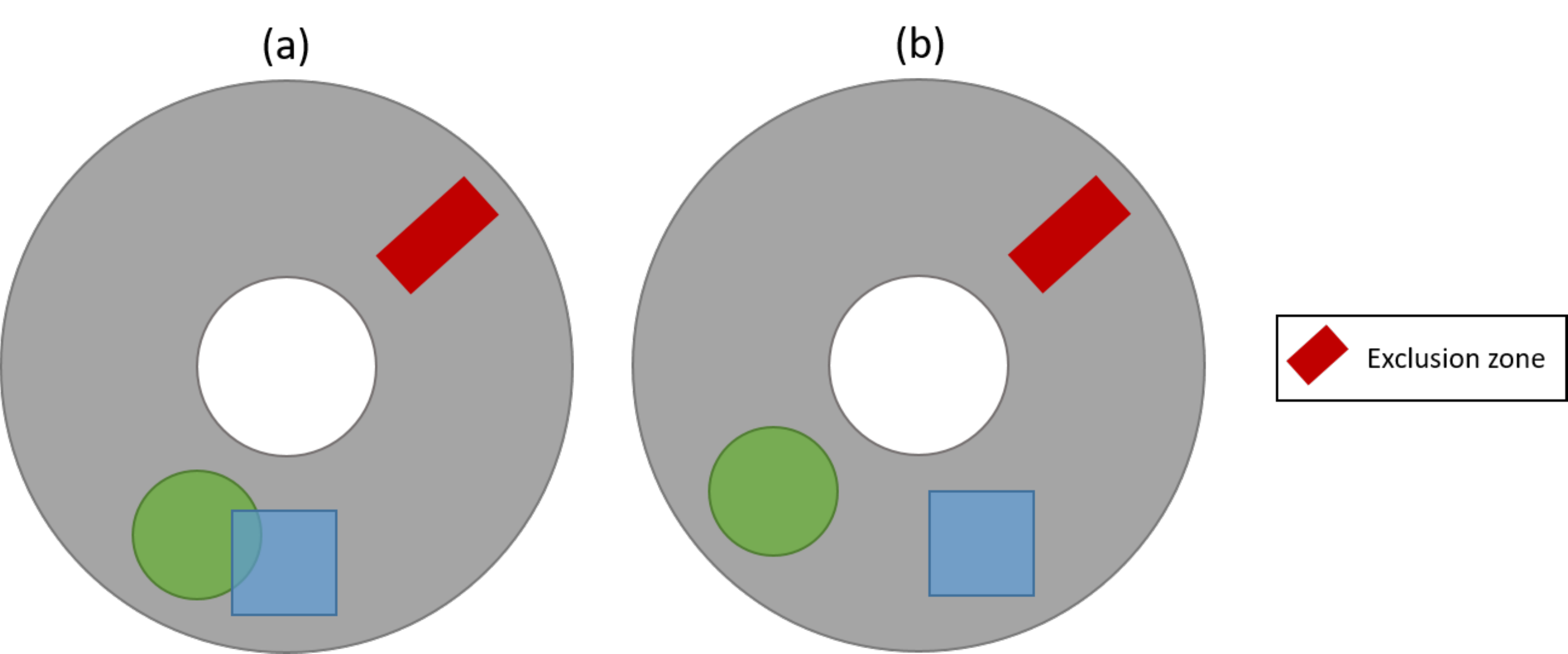}
    \caption{(a) Overlap between components (b) The overlap constraint is solved.}
    \label{fig:overlap}
\end{figure}

The constraints considered in this paper can be split into 2 categories: geometrical constraints and functional constraints. Geometrical constraints that are considered are as follows:

- No overlap between the components. Figure \ref{fig:overlap} shows two layouts where the overlap constraint is respectively violated because of two components overlapping each other and solved because no overlap is observed. \\
With the same formalism as in Section \ref{sec:inertia} the overlap constraint is expressed as:
\begin{equation*}
    h_1(\textbf{x},\textbf{z},\textbf{w})=\sum_{i=1}^{N_k-1} \sum_{j=i+1}^{N_k} \Delta V_{ij} = 0
\end{equation*}
Where $\Delta V_{ij}$ is the volume of intersection between two components $i$ and $j$, and is function of the components centroids and their orientation.\\

\begin{figure}[ht]
    \centering
    \includegraphics[width=10cm]{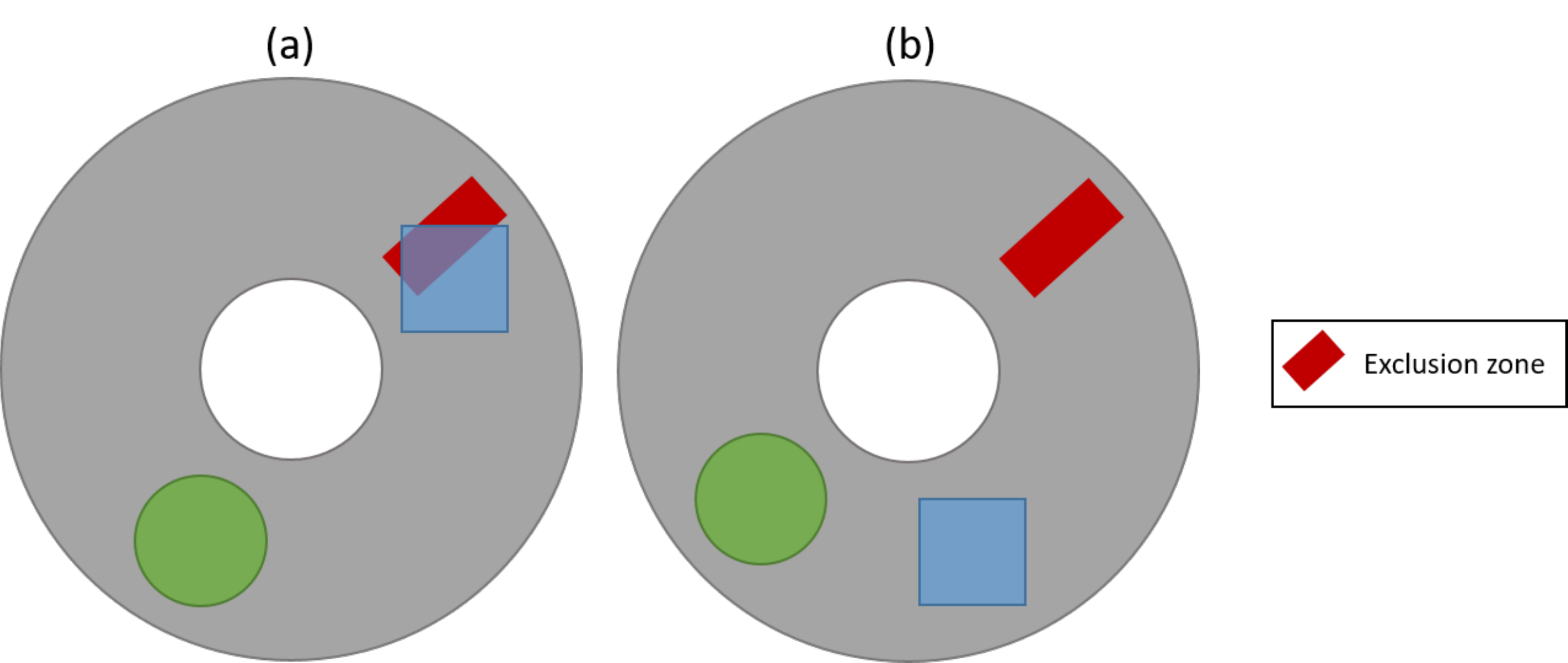}
    \caption{(a) Overlap between a component and the exclusion zone  (b) The exclusion constraint is solved.}
    \label{fig:exclusion}
\end{figure}

- No overlap between the components and the exclusion zones which may represent some components fixed to the container, for instance connection buses. Figure \ref{fig:exclusion} shows two layouts where the exclusion constraint is respectively violated because of one component overlapping the exclusion zone and its resolution.

With the same formalism as before, the exclusion constraint is expressed:
\begin{equation*}
    h_2(\textbf{x},\textbf{z},\textbf{w})= \sum_{j=1}^{N_k} \Delta V_{i}^E = 0
\end{equation*}
where $\Delta V_{i}^E$ is the volume between a component and the exclusion zone, and is function of the components and exclusion zone centroids and orientation.

\begin{figure}[ht]
    \centering
    \includegraphics[width=10cm]{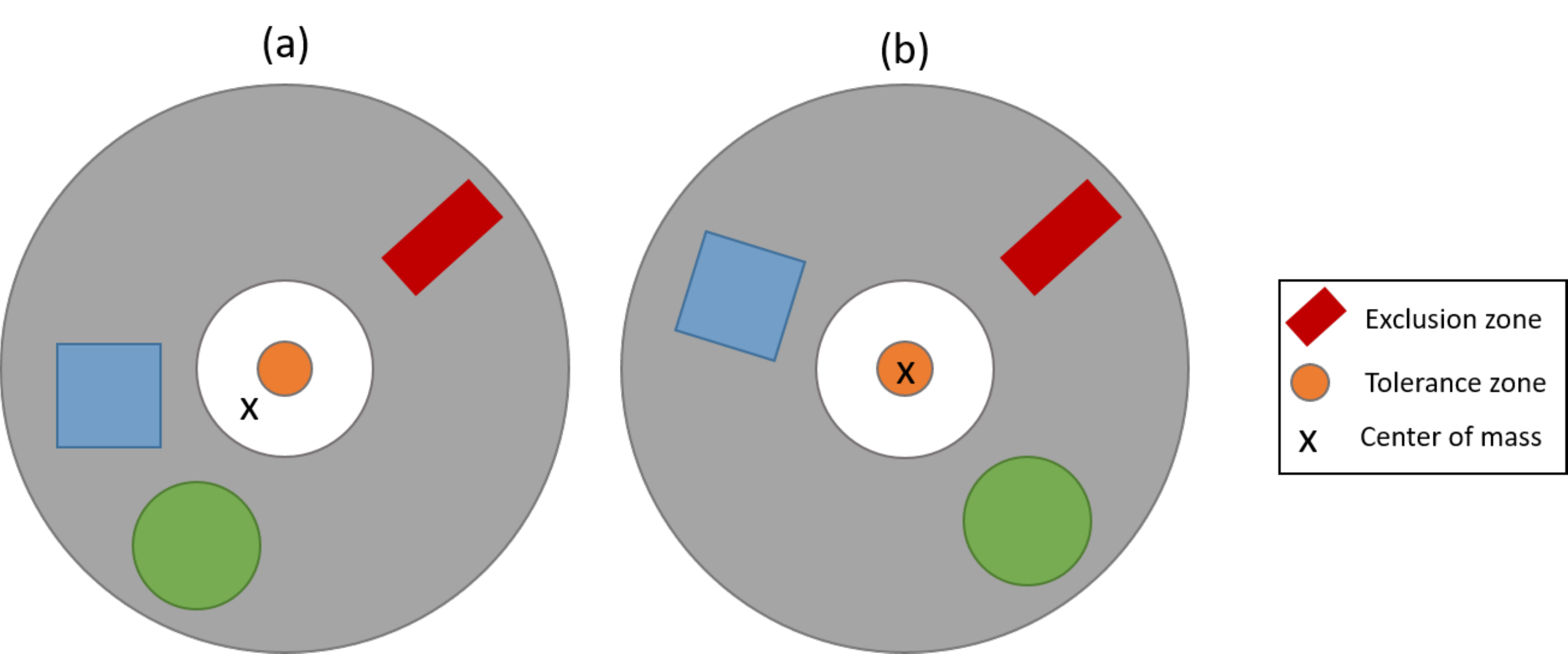}
    \caption{(a) Center of mass out of the tolerance zone (b) Center of mass in the tolerance zone.}
    \label{fig:CG}
\end{figure}

- The center of mass of the whole module must be placed at the center of the container. Figure \ref{fig:CG} illustrates two layouts where the center of mass constraint is respectively violated because the center of mass is not accurately positioned and solved because the center of mass is in the right zone. 
Using the same formalism as before, this constraint can be expressed as: 
\begin{equation*}
    g_1(\textbf{x,z,w}) = \sqrt{(x_c-x_e)^2+(y_c-y_e)^2+(z_c-z_e)^2} \leq \delta
\end{equation*}
where ($x_c,y_c,z_c$) are the coordinates of the current centroid of the whole module and ($x_e,y_e,z_e$) are the coordinates of the ideal position of the centroid. The geometric center ($0,0,0$) is considered here. 
$\delta$ represents a tolerance which corresponds, in this paper, to a circle centered about the origin of the $(Oxyz)$ coordinates system and whose radius is taken as 1\% of the outer radius of the container, without loss of generality.

\noindent A variety of functional constraints can also be added to the problem formulation via the use of special components that induce further restrictions on the feasibility of the layout.

\begin{figure}[ht]
    \centering
    \includegraphics[width=10cm]{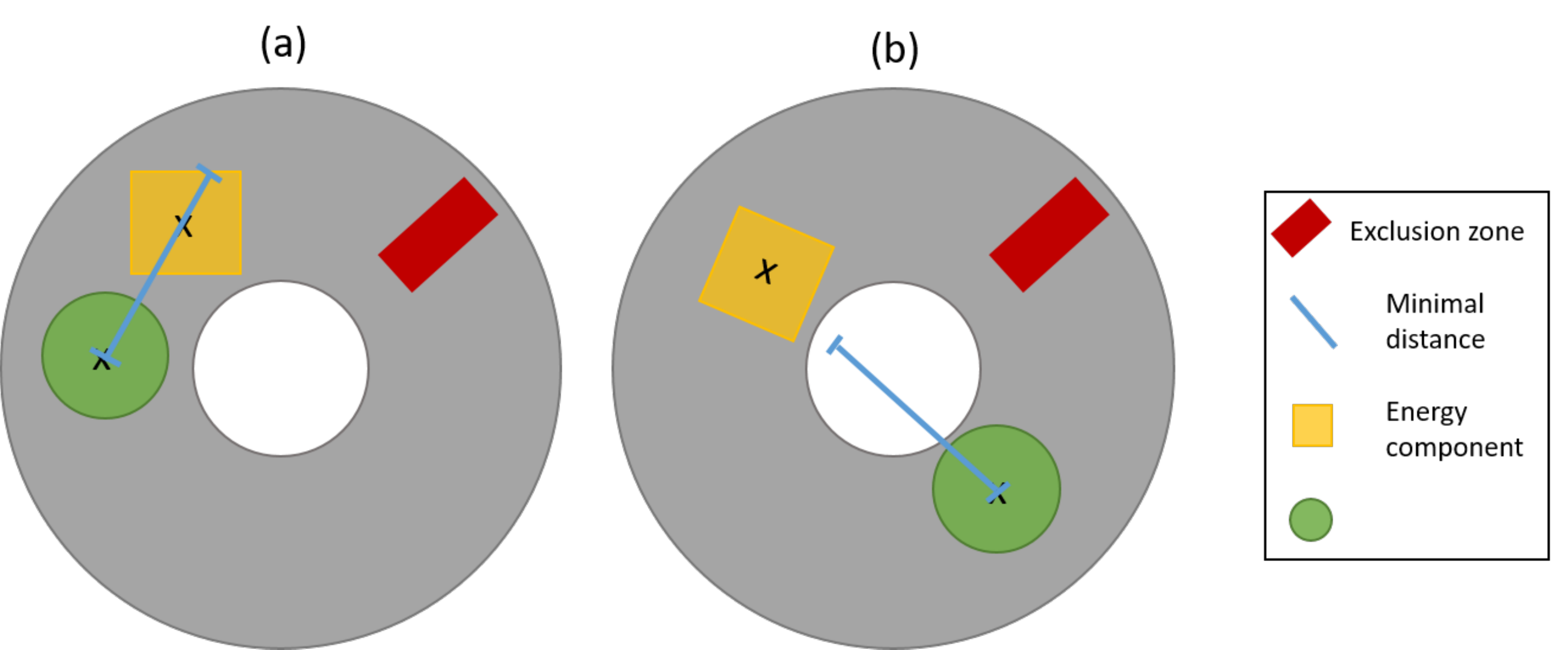}
    \caption{(a) Components too close (b) Components spaced from the minimal distance.}
    \label{fig:funccst}
\end{figure}

- For instance, the existence of specific components may involve some restrictions in the layout. For example, fuel components and energy components must be kept at a certain distance from each other for functional reasons such as not exceeding a radiation (\textit{e.g.} heat, electromagnetic) threshold at any point of the module. Figure \ref{fig:funccst} shows two layouts where the functional constraint is respectively violated because the energy and fuel components are too close and solved because the minimal distance between them is respected. With the same formalism as before, considering the configuration $k$ with $N_k$ components and $N_k^f$ fuel components and $N_k^e$ energy components, the constraint is expressed as:
\begin{equation*}
    g_3(\textbf{x},\textbf{z},\textbf{w})= \sum_{i=1}^{N_k^e} \sum_{j=1}^{N_k^f} \max(D_{min}-D(C_i^e,C_j^f),0)
\end{equation*}
Where $C_i^e$ is the energy component $i$ and $C_j^f$ is the fuel component $j$ and  $D(C_i^e,C_j^f)$ is the distance between their centroids in the current layout. $D_{min}$ is the minimal distance to be respected and is function of the type of components. Without loss of generality, other types of components can be considered within this formalism (\textit{e.g.} for electromagnetic compatibility).

\subsection{Mathematical formulation of the general variable-size optimal layout Problem}

The generic single-objective VSDS optimal layout problem can be mathematically formulated as follows:
$$
\left\{
\begin{array}{llll}
\text{Minimize } f(\textbf{x},\textbf{z},\textbf{w}) \\
\text{where: } \textbf{x} \in F_x(\textbf{w}) \subseteq \mathbb{R}^{n_{x(\textbf{w})}}, \textbf{z} \in F_z(\textbf{w}) \subseteq \mathbb{Z}^{n_{z(\textbf{w})}}, \textbf{w} \in F_w  \\
\text{subject to :}\\
\textbf{h(x,z,w)} = 0 \\
\textbf{g(x,z,w)} \leq 0
\end{array}
\right.
$$

Where $f$ is the objective function, \textbf{x} is a $n_x(\textbf{w})$-dimensional vector (continuous variables), \textbf{z} is a $n_z(\textbf{w})$-dimensional vector (discrete variables) and \textbf{w} is a $n_w$-dimensional vector (dimensional variables). \textbf{h} are the equality constraints defined by a $n_h(\textbf{w})$-dimensional vector and \textbf{g} are the inequality constraints defined by a $n_g(\textbf{w})$-dimensional vector. In this formulation, because of the dimensional variables, the domains of definition $(F_x(\textbf{w}),F_z(\textbf{w}))$ of the optimization variables are functions of the value taken by the dimensional variables. Furthermore, the number and type of constraints evolve also according to the value taken by the dimensional variables.

To summarize, the problem to solve is a large scale, mixed-variable and VSDS problem, as well as strongly constrained. The sizes of the definition spaces of the continuous and discrete variables as well as the expression of the constraints vary along with the number of components selected to be in the container. For instance, if one component is added to the list of the components, the number of continuous variables increases by 2 if it is a cylinder and by 3 if it is a cuboid (position of the center of inertia and orientation), the number of discrete variables increases by 1 (the plate to position the new component). Moreover, the expressions of the constraints will also be modified. The expressions of the overlap constraint and the exclusion constraint are changing because the number of components $N_k$ is modified. The functional constraint is changing too if the component is a fuel or an energy component. Finally, the center of mass of the module is recalculated and will be different but the expression of the centroid constraint remains unchanged.

\section{Hidden-variables mechanisms for genetic algorithms}
\label{sec:algo}
\subsection{General description of hidden-variables technique}
\label{sec:algogen}

In order to solve VSDS problems, an adaptation of an hidden-variables mechanism is proposed in this paper and is added to a GA. This mechanism consists in modifying the encoding of the design variables. Indeed, all the design variables are implemented so that all the individuals (\textit{e.g} the chromosomes) in the population have the same length. However, some of the variables are hidden, meaning that they will not be taken into account during the objective function and constraints evaluations. For the aforementioned optimal layout problem, the hidden variables represent the design variables \textbf{x} and \textbf{z} of the components of the subdivisions that are not selected by the dimensional variables \textbf{w} to be part of the current layout. On the contrary, the revealed design variables correspond to the variables of components selected to be in the container and so used to calculate the characteristics of the module such as the inertia or the position of the center of mass.

Thus, this method allows to tackle the VSDS aspect of optimization problems while ensuring that all chromosomes (corresponding to the individuals of a population) have the same length which enables to use many of existing operators. This method is preferred to the other methods mentioned in Section \ref{sec:stateV} such as Bayesian Optimization or DSMPGA. Indeed, it is well suited for the large number of design variables. It is a flexible and generic mechanism limiting the introduction of expert knowledge during the implementation process unlike the Structured-Chromosome solution and relatively simple to be implemented. In addition, it allows to keep an entire freedom of choice for the evolutionary operators unlike some techniques which do not guarantee the same number of design variables for all individuals of the population.

Depending on the method used to hide the variables, some of the evolutionary operators must nevertheless be adapted. Two methods are proposed in order to hide variables and those methods will be applied to a GA. Those methods as well as the resulting adaptations are detailed in the following section.

\subsection{Implementations of the hidden-variables mechanism with a genetic algorithm}
Different methods exist for the implementation of the hidden-variables mechanism. Two of them are described in the following section.

\subsubsection{First method: dimensional variables}
\label{sec:firstmeth}
\begin{figure}[ht]
    \centering
    \includegraphics[width=15cm]{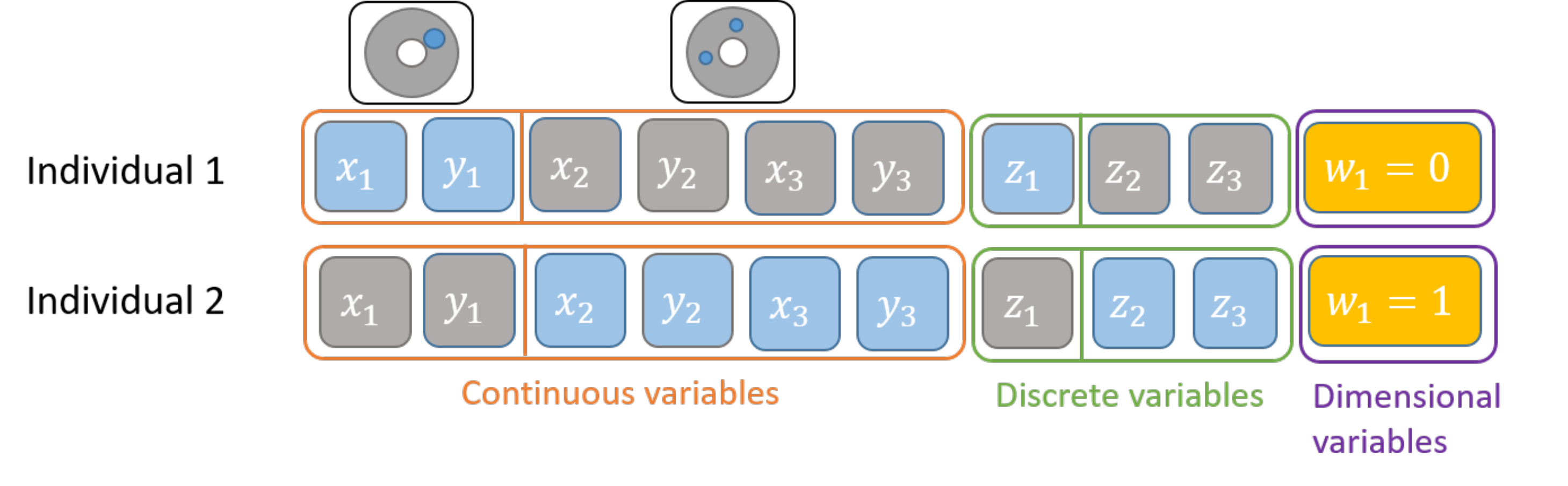}
    \caption{Illustration of the implementation of the hidden-variables mechanism thanks to dimensional variables. Blue variables are the revealed variables while grey ones are hidden ones.}
    \label{fig:dimvar}
\end{figure}

The first possibility to hide variables is to implement dimensional variables as design variables. More specifically, dimensional variables responsible for the choice of hidden or revealed variables are added to the design variables. Before evaluating the objective function and the constraints, the dimensional variables are read for each candidate solution and their values indicate which design variables of the candidate solution must be taken into account (and so which ones must be ignored) to calculate the objective and constraints.
Figure \ref{fig:dimvar} illustrates this method on an example. Taking the case of the optimal layout of a satellite module as a representative example, two subdivisions for a unique component can be chosen between one cylindrical component or two smaller ones. As only two configurations of the satellite module are considered, a unique dimensional variable taking binary values is enough to describe this feature. If the dimensional variable value is 0 then the first subdivision is chosen and the design variables linked are revealed and, on the contrary, if the dimensional variable value is 1 the second subdivision is chosen and the other design variables are revealed. It is worth noting that there exists various ways to define dimensional variables in order to describe the possible configurations. \\

Taking into account a GA as optimization algorithm, this consists in considering additional genes to handle the dimensional variables without modifying the main mechanisms and operators of the GA. Table \ref{tab:opga} sums up the operators that can be used without loss of generality.

\begin{table}[ht]
    \centering
    \begin{tabular}{|c|c|}
        \hline
         Step & Operator   \\
         \hline
         \hline
         Constraints handling & Constraint-Dominance \cite{deb2000} \\
         \hline
         Selection & Tournament  \\
         \hline
         Crossover & Simulated Binary Crossover \cite{deb} \\
         \hline
         Mutation & Polynomial Mutation \cite{talbi} \\
         \hline
         Replacement & Non-dominated truncating   \\
         \hline
         \end{tabular}
    \caption{Operators used for GA with hidden-variables mechanism}
    \label{tab:opga}
\end{table}

\subsubsection{Second method: tags}
\label{sec:secondmeth}
\begin{figure}[ht]
\begin{subfigure}{.5\textwidth}
  \centering
  \includegraphics[width=.8\linewidth]{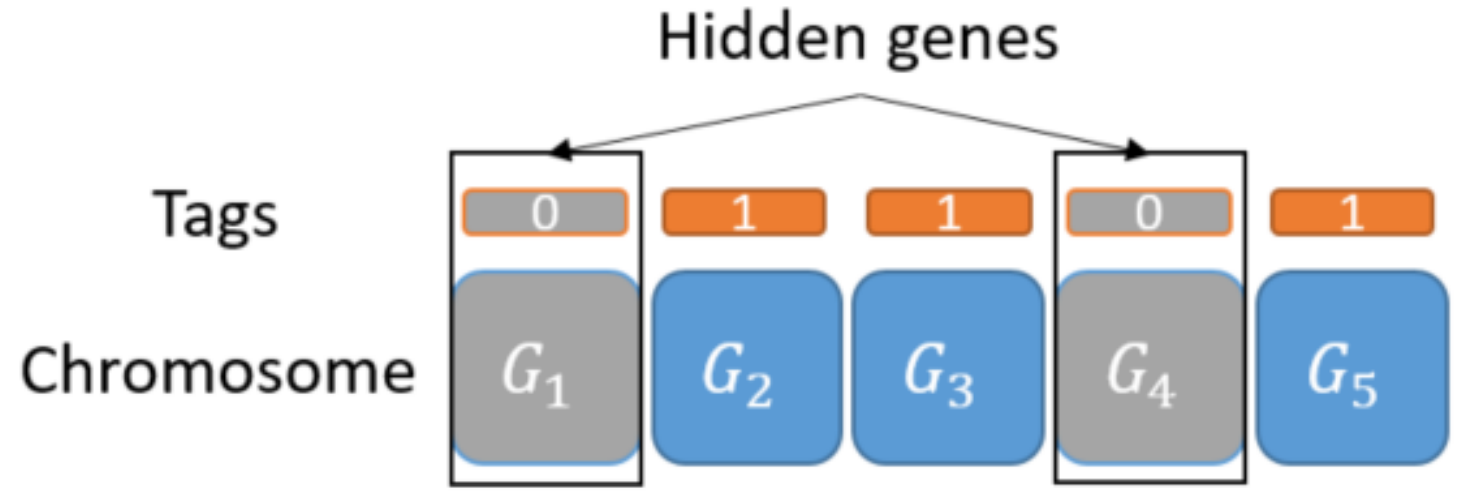}  
  \caption{Definition of the tags.}
  \label{fig:tag1}
\end{subfigure}
\begin{subfigure}{.5\textwidth}
  \centering
  \includegraphics[width=1\linewidth]{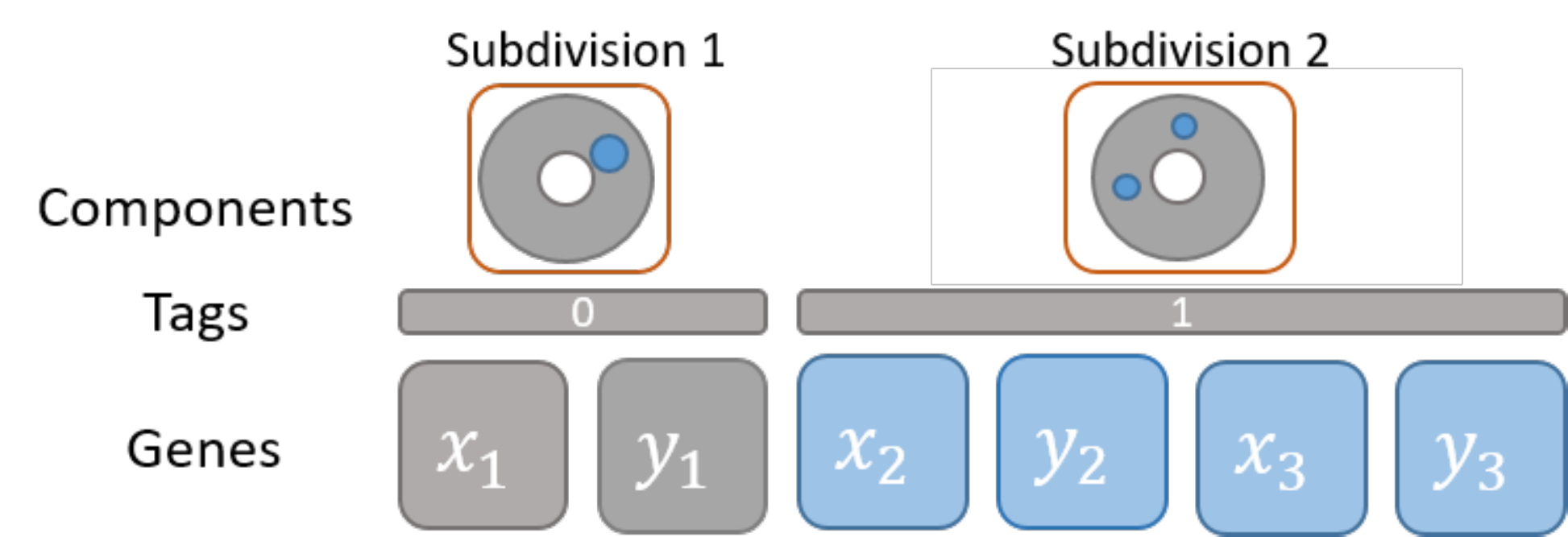}  
  \caption{Adaptation of the tag method.}
  \label{fig:tag2}
\end{subfigure}
\caption{Illustration of the tag mechanism.}
\label{fig:tag}
\end{figure}

Another possible way to deal with VSDS is to introduce tags as proposed in \cite{abdelkhalik2011}. In this method, a tag vector is attached to each chromosome of the population such that a tag value is assigned to each gene of the chromosome. If the value of the tag is $0$, then the corresponding gene will not participate in the objective function evaluation and reciprocally. This mechanism is shown in  Figure \ref{fig:tag1}. The tags are a direct translation of the dimensional variables: their role is to choose which variables will be expressed or not during the fitness evaluation as well as the corresponding constraints to activate. Figure \ref{fig:tag2} illustrates the adaptation on the same example as in the previous section. If the second subdivision of the component is chosen to be part of the layout, the tag vector will activate all the design variables (coordinates of the center of mass and orientation if needed) to be part of the objective function  and constraints evaluations. 

As the chromosomes are enhanced by tags, the operators of the GA must be compatible with the tags which will take part in the evolution process. The chromosomes and tags will undergo independent operators of crossover and mutation as proposed in \cite{abdelkhalik2011}.
\vspace{0.5cm}

\noindent \textbf{Crossover} \par

\begin{figure}[ht]
    \centering
    \includegraphics[width=15cm]{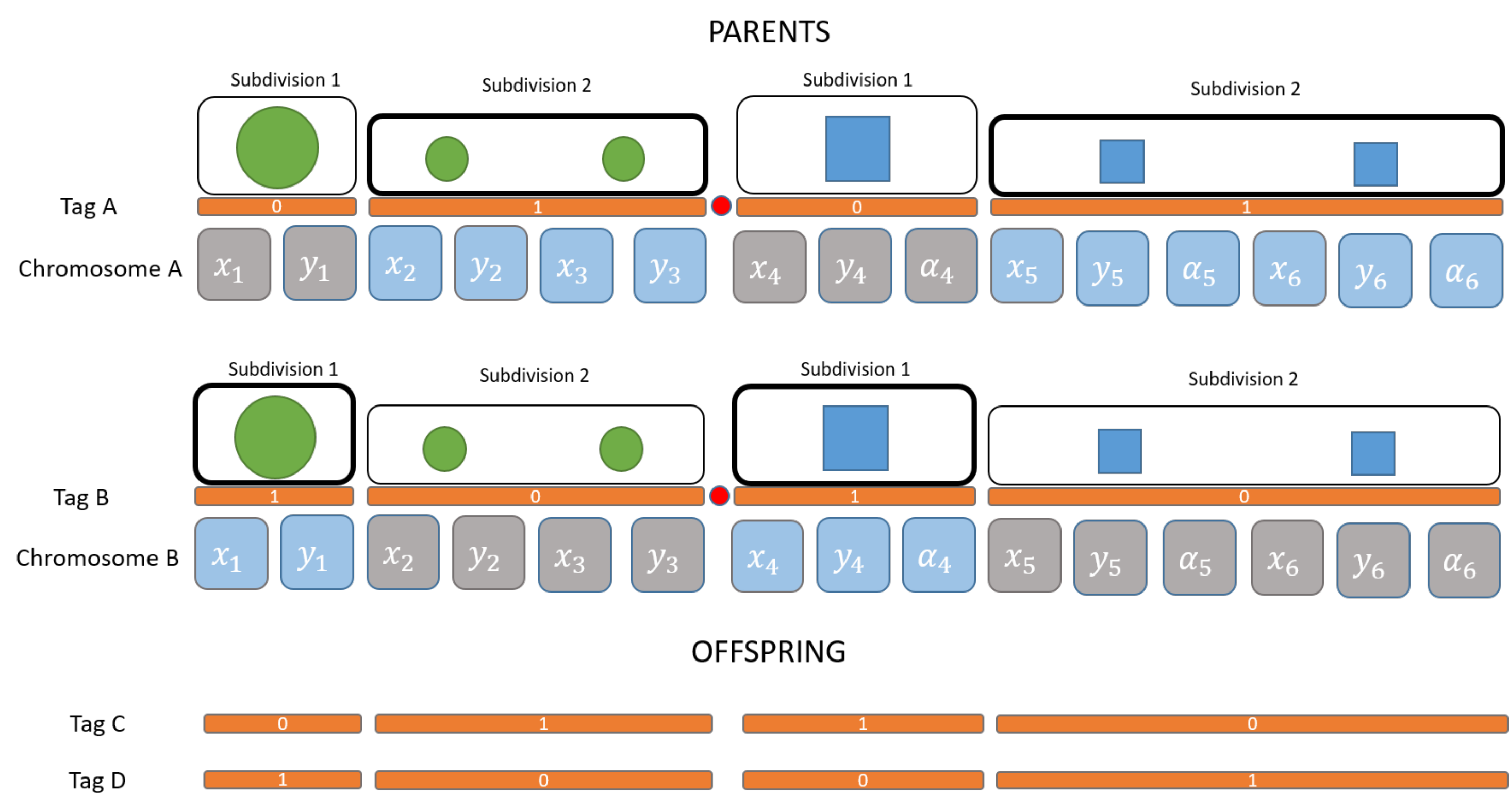}
    \caption{Crossover operator for the tags on an example.}
    \label{fig:crossovertags}
\end{figure}

For the chromosomes, the crossover operator can be chosen among all the operators available for "classical" genetic algorithms \cite{talbi}. For the tags, a \textit{n}-points crossover operator is chosen in this paper, without loss of generality \cite{talbi}. This mechanism must be adapted as not all the points of the tag vector can be crossover points because of the fact that the design variables are not independent from each other. The crossover operator on the tags must provide a feasible configuration of subdivisions. Figure \ref{fig:crossovertags} illustrates the principle of the \textit{n}-points crossover on the tags adapted for the subdivisions of the components. On the example, two components able to subdivide themselves into two smaller components are considered so that there are four possible configurations of components in the container. For each component, only one of the two subdivisions can be present in the container and so in the tag vector, amongst the three points that may be considered as crossover points only one (indicated in red) guarantees the fact that the crossover operation will lead to feasible individuals. 

\vspace{20pt}

\noindent \textbf{Mutation}\par
\begin{figure}[ht]
    \centering
    \includegraphics[width=15cm]{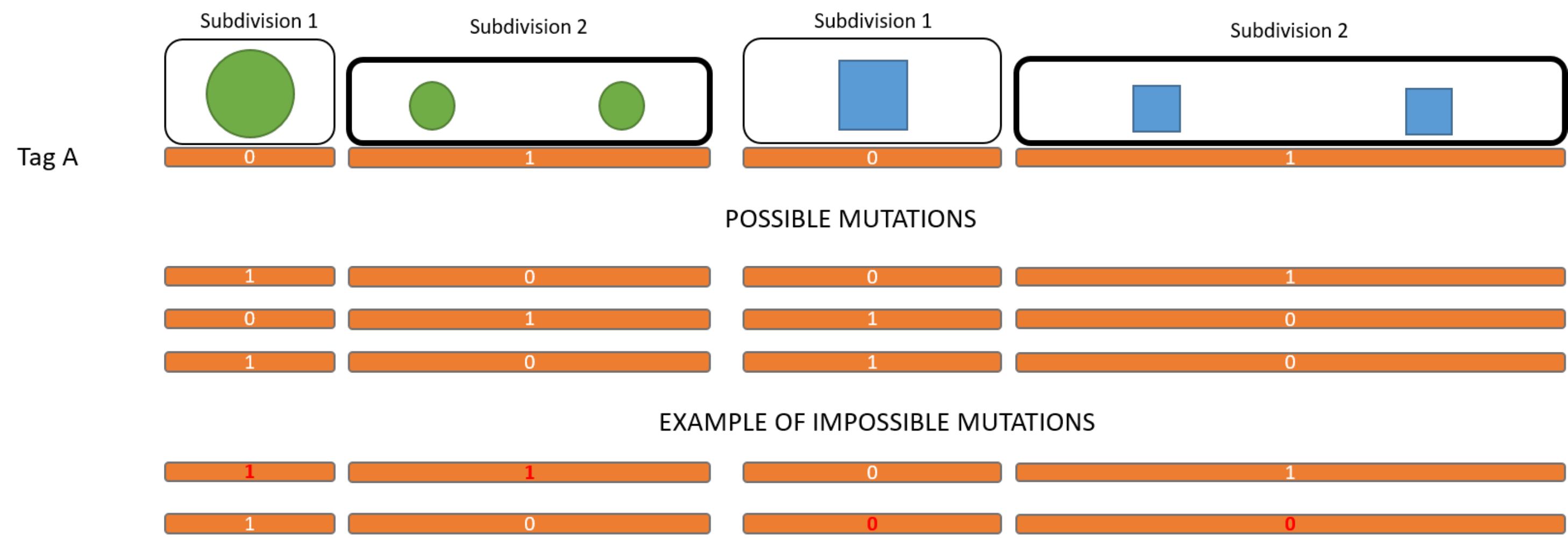}
    \caption{Mutation operator for the tags on an example.}
    \label{fig:mutationtags}
\end{figure}

For the mutation operator, again the mechanism differs according to the chromosomes or tags. For the chromosomes, the mutation operator can be chosen amongst all the operators available for "classical" genetic algorithms. For the tags, a bit flip operator can be applied \cite{talbi}. Like the crossover operator, it must be adapted for the subdivision case to provide feasible solutions. The bit flip consists in fact in flipping the subdivision chosen. Hence the tags cannot be flipped randomly as for each component only one tag must be equal to one in order to select a unique subdivision amongst the possible ones. Figure \ref{fig:mutationtags} illustrates the mutation operator designed for the tags. Only a few mutations are possible to guarantee the feasibility of the solution. 

As for the crossover operators, the probability of mutation for chromosomes and tags can differ as a 2-dimensional mutation operator is also used. \par

The two ways described in this section to implement the hidden-variables mechanism allowing a GA to tackle the VSDS nature of the problem (dimensional variables and tags) will  be compared in Section \ref{sec:toy} in a toy case and in Section \ref{sec:sat} on a realistic design problem.

\section{Illustration on a toy case}
\label{sec:toy}
In this section, a test case that can be analytically solved is presented. This example allows to validate the hidden-variables mechanisms  described in Section \ref{sec:algo}. Indeed, the methods are implemented on a very simple case in order to illustrate the hidden-variable mechanisms on a first example, as well as to illustrate the ability of the adapted algorithms to reach the optimum obtained analytically.

\subsection{Configuration}
\label{sec:bench}
The configuration of the test case is as follows:
\begin{itemize}
    \item \textbf{Container:} one single-sided bearing plate.
    \item \textbf{Components:} two cylindrical components which can be subdivided into 2 sub-components resulting in the 4 different configurations of subdivisions illustrated in Figure \ref{fig:configtoy}. For illustration purposes, in all the following figures, one of the components will be colored in blue while the other in green to differentiate them.
    \item \textbf{Design variables:} the coordinates of the center of gravity of each component parameterized in cylindrical coordinates (\textit{e.g.} radius and angle: $(r_{j,k}^i,\theta_{j,k}^i)$ for component $i$, sub-component $k$ of the subdivision $j$). They correspond to 12 genes in each chromosome.
    \item \textbf{Objective function:} the inertia of the module (to be minimized).
    \item \textbf{Constraints:} no overlapping between the components and the center of gravity must be placed at the center of the bearing plate (with a tolerance zone set equal to 1\% of the outer radius of the plate).
\end{itemize}

\begin{figure}[ht]
    \centering
    \includegraphics[width=15cm]{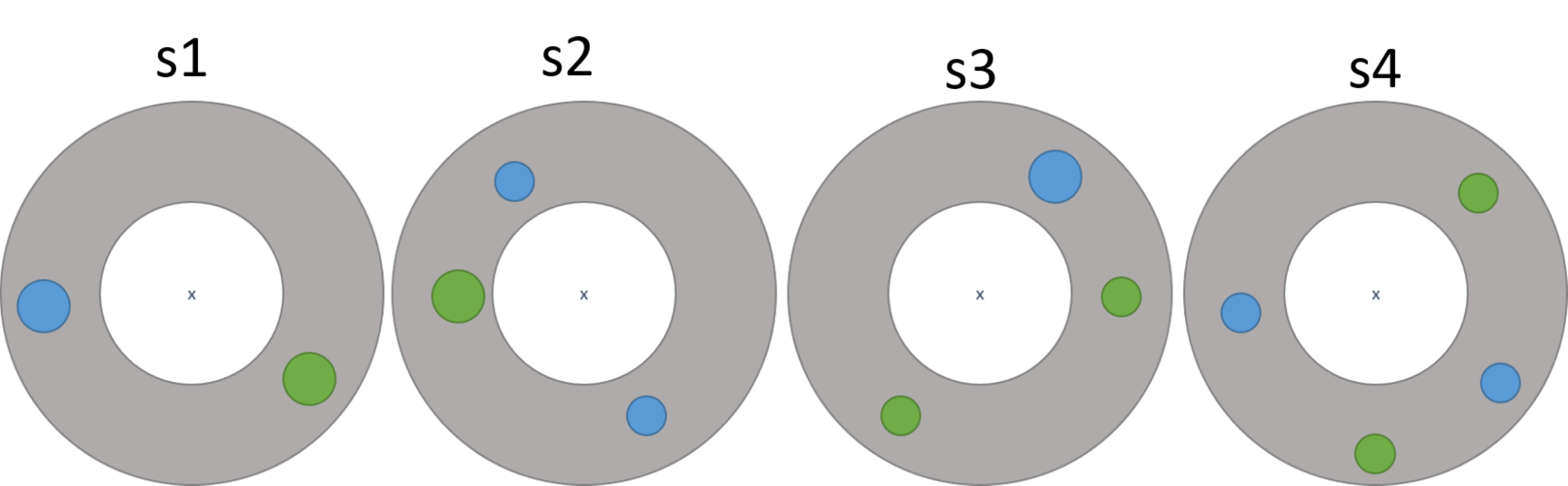}
    \caption{The four possible configurations of subdivisions for the toy case.}
    \label{fig:configtoy}
\end{figure}

\begin{figure}[ht]
    \centering
    \includegraphics[width=7cm]{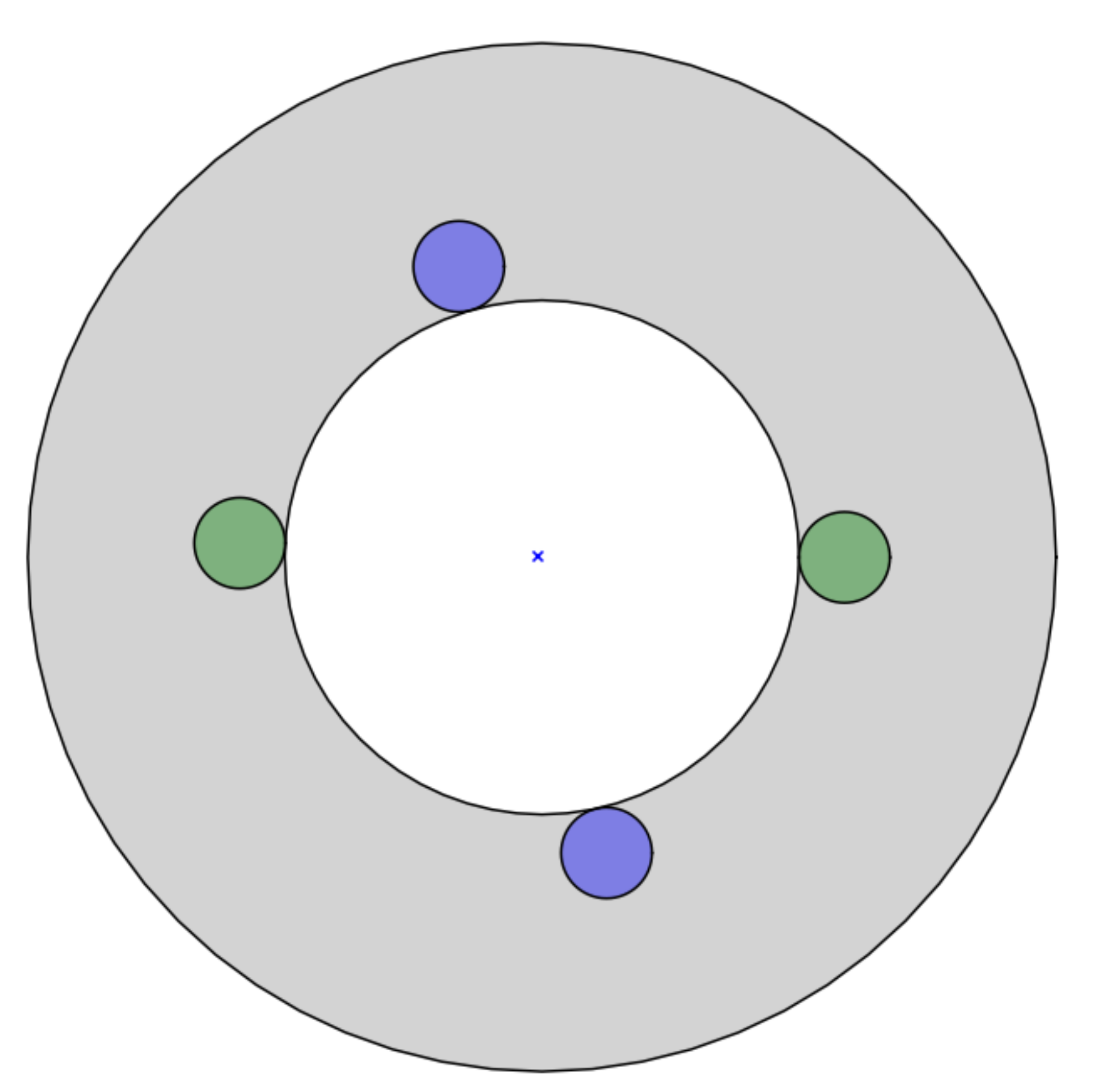}
    \caption{One of the optimum configurations and layouts of the case test}
    \label{fig:optimtest}
\end{figure}

Figure \ref{fig:optimtest} shows one of the optimal layouts (the problem is rotationally symmetric and so there is an infinite number of optimal solutions). For this toy case, the optimal value of inertia, that can be found analytically, is 4.49$\times 10^5$.

\subsection{Hidden-variables mechanism implementations}
\begin{figure}[ht]
    \centering
    \includegraphics[width=16cm]{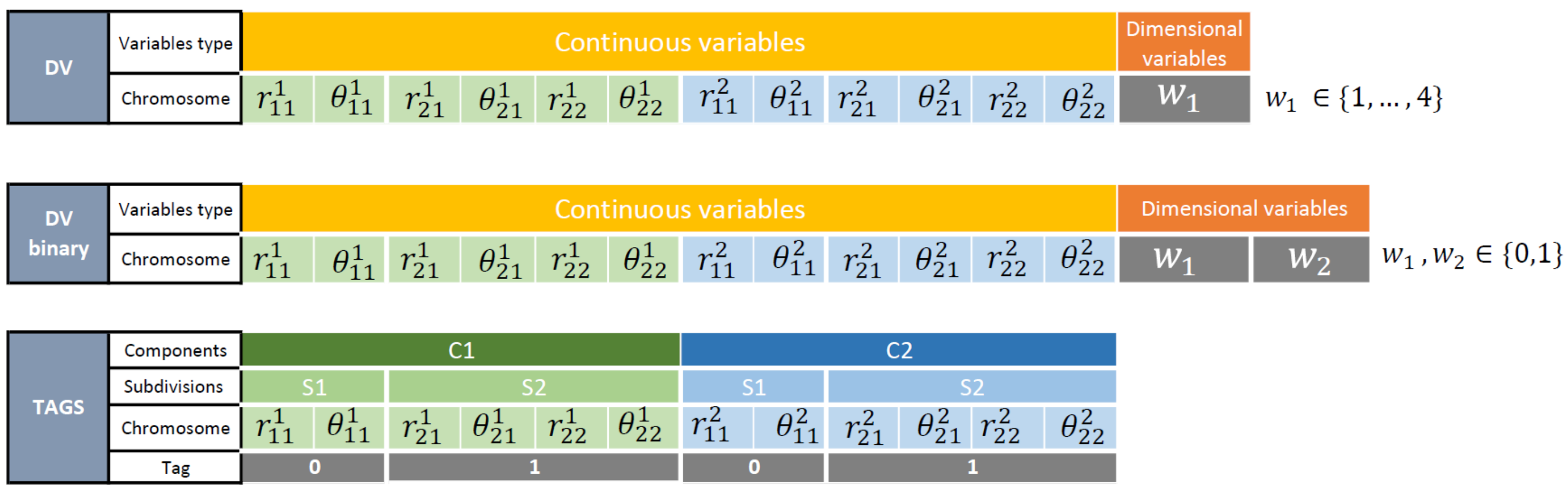}
    \caption{Three ways of implementing the design variables.}
    \label{fig:chrtest}
\end{figure}

As described in Section \ref{sec:algo}, different ways for implementing the hidden-variables mechanism are possible. Three implementations will be studied on the toy case in order to verify their behavior. Figure \ref{fig:chrtest} illustrates the three corresponding encodings of the chromosomes.

The two first ones correspond to the first method developed in Section \ref{sec:firstmeth} which uses dimensional variables (DV) directly integrated as design variables. Here, either one dimensional variable is used taking a value between 1 and 4 or two dimensional variables are used taking binary values. In both cases, the dimensional variables represent the 4 different configurations and their values allow to express the corresponding genes during the objective and constraint evaluations. Those three ways to implement the proposed mechanism will be applied to a GA.

To sum up, the three methods implemented on the toy case are:
\begin{itemize}
    \item A hidden-variables GA with the tag method to hide genes (GA-tags) ;
    \item A GA with the numerical dimensional variables method as hidden-variables mechanism (GA-DV-num) ;
    \item A GA with the binary dimensional variables method as hidden-variables mechanism (GA-DV-bin).
\end{itemize}

\begin{figure}[ht]
    \centering
    \includegraphics[width=10cm]{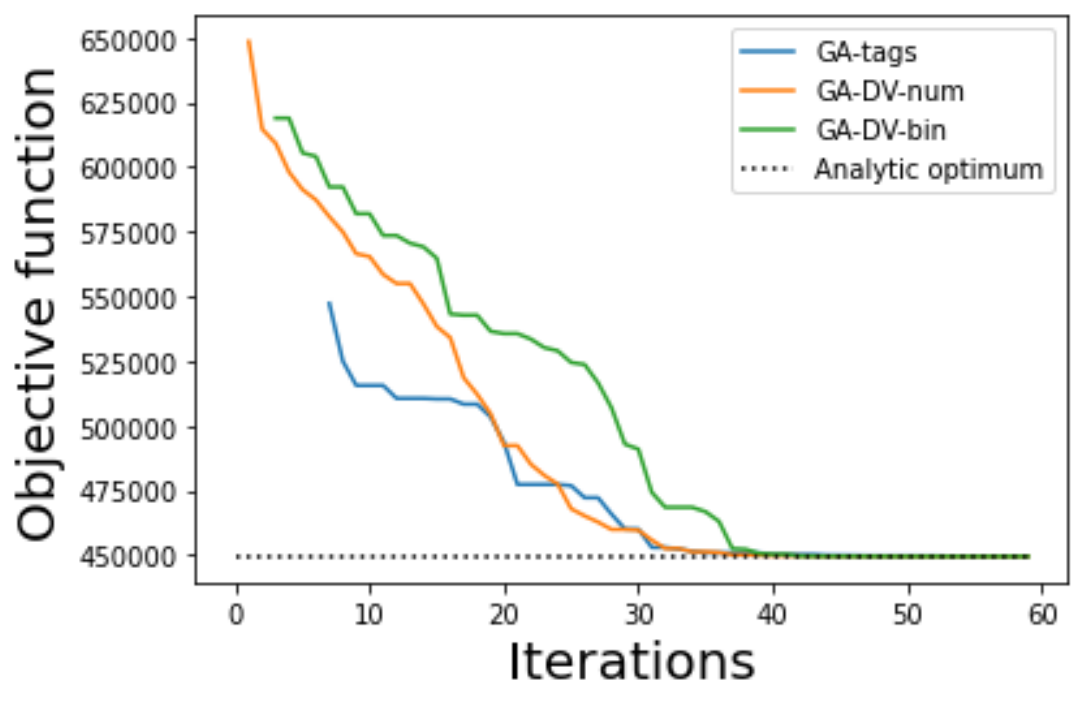}
    \caption{Convergence curves for each hidden-variables mechanism implementations with the described GA.}
    \label{fig:besttest}
\end{figure}

\ref{sec:appendix_test} encompasses the Tables which sum up the operators, strategies and hyper-parameters (set thanks to a parametric study) for all algorithms: GA-tags on Table \ref{tab:gatagstestt}, as well as GA-DV-bin and GA-DV-num on Table \ref{tab:gadvtest}.
Table \ref{tab:paramtc} in \ref{sec:appendix_test} sums up the numerical configuration for all three algorithms.
Figure \ref{fig:besttest} shows the convergence curves obtained for each method.

The results show that the GA enhanced by several hidden-variables mechanisms implementations manage to reach the optimum analytically obtained of the toy case. On this example, all the three methods are quite equivalent in terms on convergence speed. Indeed, these methods require about 40 iterations to find the optimal configuration.

Thus, this toy case shows that a GA, coupled with an appropriate genotype representation, is indeed capable of successfully solving a VSDS problem. In the next section, the various methods are being applied to a more representative application to test their efficiency on a more complex test case.

\newpage
\section{Application of the proposed hidden-variables mechanisms to the optimal layout of a satellite}
\label{sec:sat}

In order to illustrate the performance of the proposed methods on a representative industrial case, the application of the satellite layout provided from \cite{liu,cui2019,wang2009} is extended as a VSDS problem. The configuration of the problem and numerical schemes are outlined and the obtained results analyzed. 

\subsection{Configuration of the problem}
\label{sec:appli}
Considering the INTELSAT-III satellite introduced in Section \ref{sec:intro}, some changes are added as well as the variable-size design space aspect. First of all, the orientations of the components are now continuous design variables instead of discrete ones. An exclusion zone is added on the bearing plate in order to break the symmetry of the problem. A functional constraint is also added in order to simulate the inconsistency between some components. All these modifications make the problem solving much complicate. 
To sum up, the configuration of the application case is as follows:
\begin{itemize}
    \item \textbf{Container:} one one-sided plate of the satellite module with an exclusion zone.
    \item \textbf{Components:} twelve components (cylinders and cuboids) which can be subdivided into 2 to 4 subcomponents such that there are 3888 possible configurations as detailed in Table \ref{tab:decompo} in \ref{sec:appendix0}.
    \item \textbf{Design variables:} the 2D position of the center of gravity of each component and its orientation (as a continuous variable) so 129 genes.
    \item \textbf{Objective function:} the inertia of the module (to be minimized).
    \item \textbf{Geometric constraints:} no overlapping between the components. No overlapping between the components and the exclusion zones. The center of gravity must be placed at the center of the bearing plate (with a tolerance of 1\% of the outer radius of the plate).
    \item \textbf{Functional constraint:} the fuel components centroids must be at a minimal distance of 300 mm from the energy components centroids. 
\end{itemize}

Thus, the length of the chromosomes for this configuration varies linearly along with the number of cylinder and cuboid components. If $N_{cyl}$ and $N_{cub}$ corresponds respectively to the number of cylinders and cuboids which can possibly be chosen to be part of the layout, the length of the chromosomes $L_c$ can be expressed as follows:

\begin{equation}
    L_c = 2N_{cyl}+3N_{cub}
\end{equation}

Moreover, the size of the components will be modified in order to study the performance of the proposed algorithm with respect to the occupation rate of the container. \\
The occupation rate (OR) is defined as:

\begin{equation}
    OR=\frac{A_{components}}{A_{container}}
\end{equation}
where $A_{components}$ and $A_{container}$ correspond respectively to the total area of components and to the area of the available space in the container.

\begin{figure}[ht]
    \centering
    \includegraphics[width=15cm]{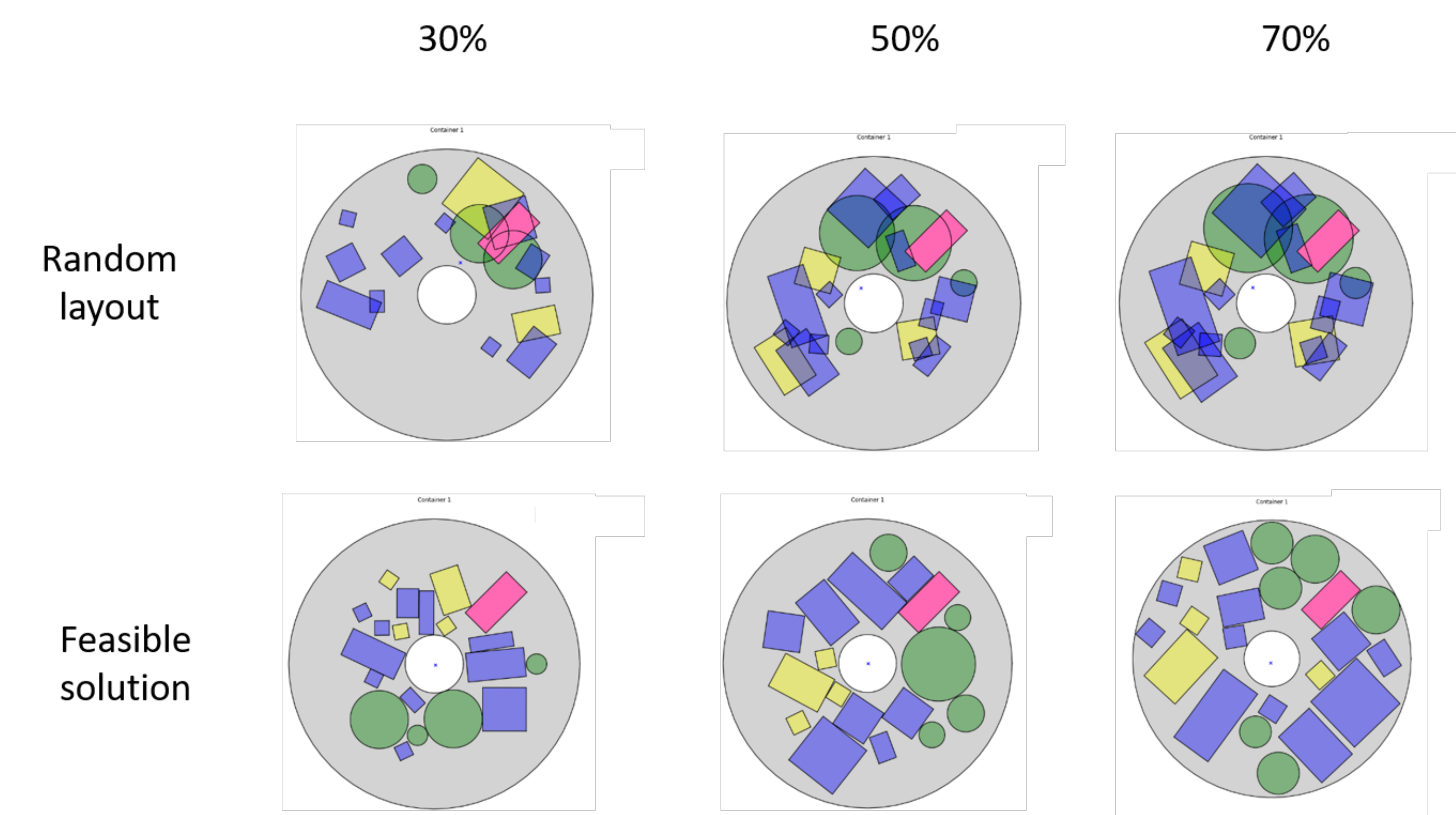}
    \caption{Evolution between the initial plate and a feasible layout for the 3 occupation rates. The pink rectangle corresponds to the exclusion zone, the green cylinders correspond to the fuel components, the yellow cuboids to the energy components and the blue cuboids to other components (\textit{e.g.} electronics ...).} 
    \label{fig:compact}
\end{figure}

In this section, the satellite module problem described in Section \ref{sec:appli} is derived for five occupation rates of the container: 30\%, 40\%, 50\%, 60\% and 70\%.
For summary purpose, Figure \ref{fig:compact} shows the evolution on the bearing plate between a first random layout and one corresponding solved layout for 3 different occupation rates of the plate: 30\%, 50\%, and 70\%. One can observe that the occupation rate induces overlap between components and the constraints may be more and more difficult to be satisfied.

\subsection{Implementations of the hidden-variables mechanisms }
In this section, both possible implementations of the proposed mechanism are applied to a GA in order to address the satellite module layout problem. 

\subsubsection{First method: dimensional variable implementation (GA-DV)}

The first implementation of the hidden-variable mechanism consists in a GA for which chromosomes have been enhanced with dimensional variables (GA-DV), as described in Section \ref{sec:firstmeth}. Those variables are responsible for choosing the components to be placed in the container. Due to the high combinatory of the problem (3888 possible configurations), a practical way to implement this method is to extend the chromosomes by as many dimensional variables as components that can be subdivided in the problem.

Table \ref{tab:chrdv} illustrates the partial structure of a corresponding chromosome. For synthesis and understanding purposes, only one cylindrical component is considered. This component can either be placed itself in the container (S1), or be subdivided into two or three smaller components (respectively S2 or S3). The design variables correspond to the coordinates of the center of inertia of each sub-components (located in cylindrical coordinates) in the container (highlighted in orange). The chromosome is then elongated by as many dimensional variables as subdivided components. Consequently, in the example of Table \ref{tab:chrdv} only one dimensional variable (highlighted in grey) is necessary taking the values from 1 to 3. The value of the dimensional variable indicates that the two sub-components of the second subdivision are chosen to be placed in the container. Thus, only design variables $r_{21}^1, \theta_{21}^1, r_{22}^1 \text{ and } \theta_{22}^1$ of this chromosome take part into the objective and constraints evaluations.

\begin{table}[ht]
    \centering
    \begin{tabular}{|c||c|c|c|c|c|c|c|c|c|c|c|c|c|}
        \cline{1-13}
        
        \rowcolor[rgb]{0,0.6,0} Components  & \multicolumn{12}{c|}{C1}  \\
        
        \rowcolor[rgb]{0.5,1,0} Subdivisions & \multicolumn{2}{c|}{S1} & \multicolumn{4}{c|}{S2} & \multicolumn{6}{c|}{S3} \\
        \hline
        \rowcolor[rgb]{1,0.5,0} Chromosome & $r_{11}^1$ & $\theta_{11}^1$ & $\mathbf{r_{21}^1}$ & $\mathbf{\theta_{21}^1}$ &  $\mathbf{r_{22}^1}$ & $\mathbf{\theta_{22}^1}$ & $r_{31}^1$ & $\theta_{31}^1$ & $r_{32}^1$ & $\theta_{32}^1$ & $r_{33}^1$ & $\theta_{33}^1$ & \cellcolor[rgb]{0.8,0.8,0.8}{\textbf{2}} \\
        \hline
        
    \end{tabular}
    \caption{Structure (partial) of a chromosome for one cylindrical component for the GA-DV method.}
    \label{tab:chrdv}
\end{table}

\subsubsection{Second method: tags implementation (GA-tags)}

The second method to implement the hidden-variables mechanism in a GA corresponds to the tag implementation (GA-tags) described in Section \ref{sec:secondmeth}. 
Table \ref{tab:chrtag} shows the principle of chromosomes and tags implementation. The same example as in the previous section is used to illustrate the tag mechanism for the application case. Indeed, a tag vector is attached to the chromosome and aims to select one possible subdivision of the component to be part of the layout. The second subdivision is chosen thanks to the tag value equal to 1 in second place.

\begin{table}[ht]
    \centering
    \begin{tabular}{|c||c|c|c|c|c|c|c|c|c|c|c|c|}
        \hline 
        \rowcolor[rgb]{0,0.6,0} Components  & \multicolumn{12}{c|}{C1}  \\
        \hline  
        \rowcolor[rgb]{0.5,1,0} Subdivisions & \multicolumn{2}{c|}{S1} & \multicolumn{4}{c|}{S2} & \multicolumn{6}{c|}{S3} \\
        \hline
        \rowcolor[rgb]{1,0.5,0} Chromosome & $r_{11}^1$ & $\theta_{11}^1$ & $\mathbf{r_{21}^1}$ & $\mathbf{\theta_{21}^1}$ &  $\mathbf{r_{22}^1}$ & $\mathbf{\theta_{22}^1}$ & $r_{31}^1$ & $\theta_{31}^1$ & $r_{32}^1$ & $\theta_{32}^1$ & $r_{33}^1$ & $\theta_{33}^1$  \\
        \hline
        \rowcolor[rgb]{0.8,0.8,0.8} Tag & \multicolumn{2}{c|}{0} & \multicolumn{4}{c|}{\textbf{1}} & \multicolumn{6}{c|}{0} \\
        \hline 
    \end{tabular}
    \caption{Structure (partial) of a chromosome and associated tag (in grey) for one cylindrical component for the GA-tags method.}
    \label{tab:chrtag}
\end{table}

\subsubsection{Numerical settings}

Given the combinatorial dimension of the problem, different operators have been tested and compared in order to promote diversity and to prevent premature convergence. A Taguchi's experiment plan \cite{taguchi} have been adopted to analyze in depth the performance of a GA enhanced by the proposed mechanisms, with several evolutionary operators, on the satellite module layout problem and with different occupation rates. The Taguchi's experiment plan allows to study exploration and exploitation capabilities of the algorithms and therefore to select the best sequence of operators. 
Moreover, a parametric study has been run to set the hyperparameters. 
The following sequence of operators as well as hyperparameters are selected after a test campaign:

\begin{itemize}
    \item \textbf{Constraint handling}: Stochastic Ranking \cite{runarsson} with probability $P_f=0.45$.
    \item \textbf{Selection}: Tournament with 15 individuals.
    \item \textbf{Crossover design variables}: Simulated Binary Crossover (SBX) \cite{deb}, with probability $P_c^{g}=0.9$.
    \item \textbf{Crossover dimensional variables} (for GA-DV): SBX with probability $P_c^{DV}=0.9$.
    \item \textbf{Crossover tags} (for GA-tags): 1-point, with probability $P_c^{t}=0.9$.
    \item \textbf{Mutation}: Polynomial Mutation (PM) \cite{talbi} with probability $P_m=0.2$.
    \item \textbf{Replacement}: Non-Dominated Truncating \cite{deb2002}.
\end{itemize}

The GA-tags and GA-DV methods are ran 20 times for the five occupation rates. For each occupation rate, the algorithms are randomly initialized, even though those initializations are the same between the two methods so that they are comparable. The objective evaluations cost is set to $3.5\times 10^5$, distributed as follows: 500 individuals in the populations and 7000 generations.

\subsection{Experimental results}

\begin{figure}[h!]
\begin{subfigure}{.45\textwidth}
  \centering
  \includegraphics[width=\linewidth]{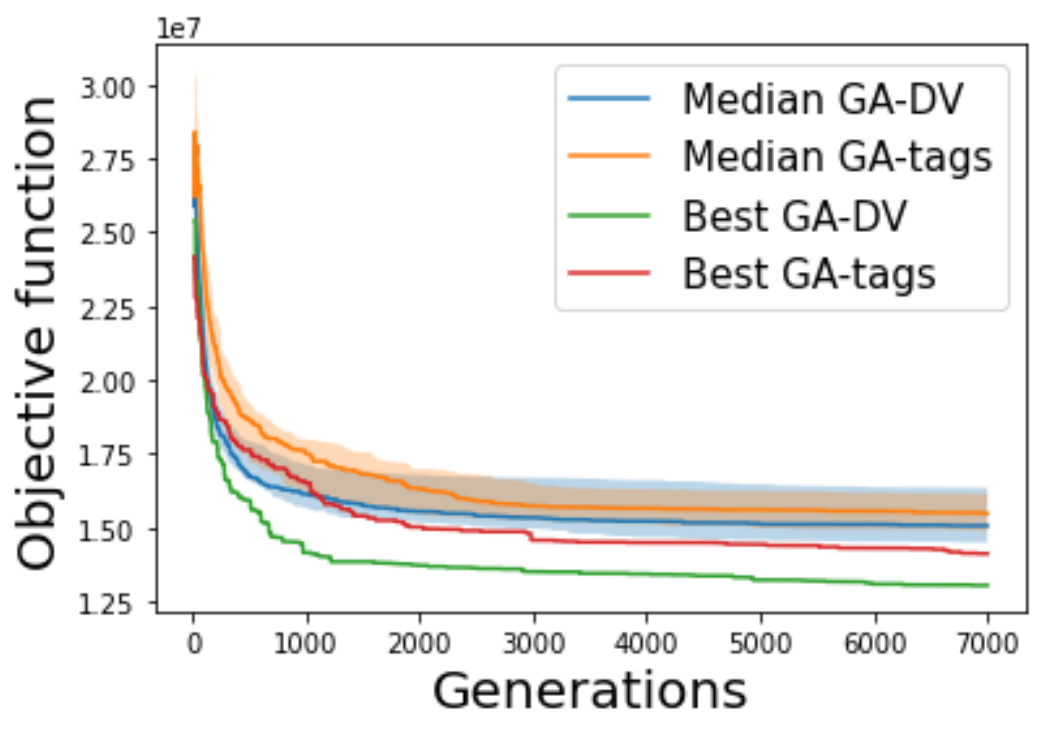}  
  \caption{Occupation rate: 30\%}
  \label{fig:mediantc}
\end{subfigure}
\begin{subfigure}{.45\textwidth}
  \centering
  \includegraphics[width=\linewidth]{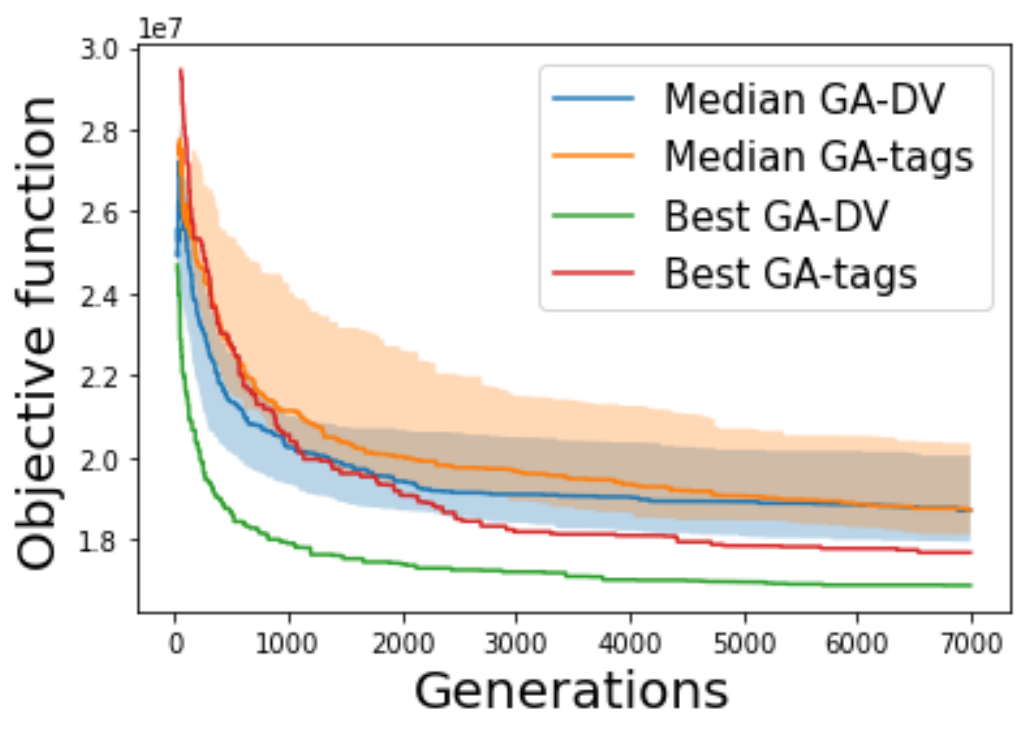} 
  \caption{Occupation rate: 40\%}
  \label{fig:met1}
\end{subfigure}
\begin{subfigure}{.45\textwidth}
  \centering
  \includegraphics[width=\linewidth]{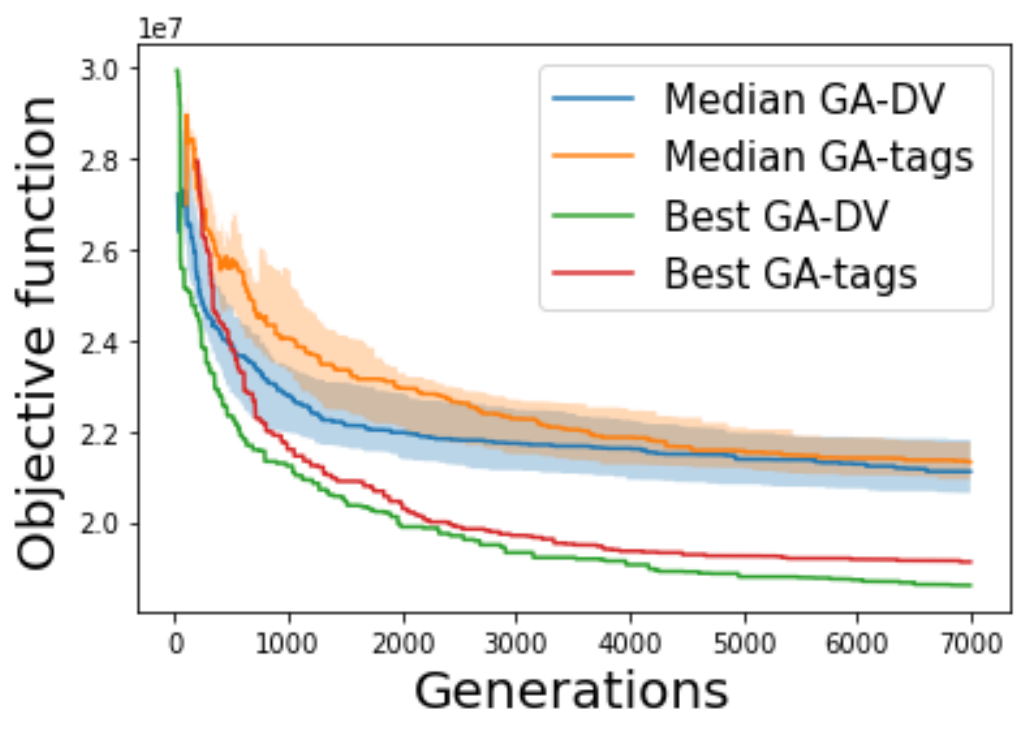} 
  \caption{Occupation rate: 50\%}
  \label{fig:met2}
\end{subfigure}
\begin{subfigure}{.45\textwidth}
  \centering
  \includegraphics[width=\linewidth]{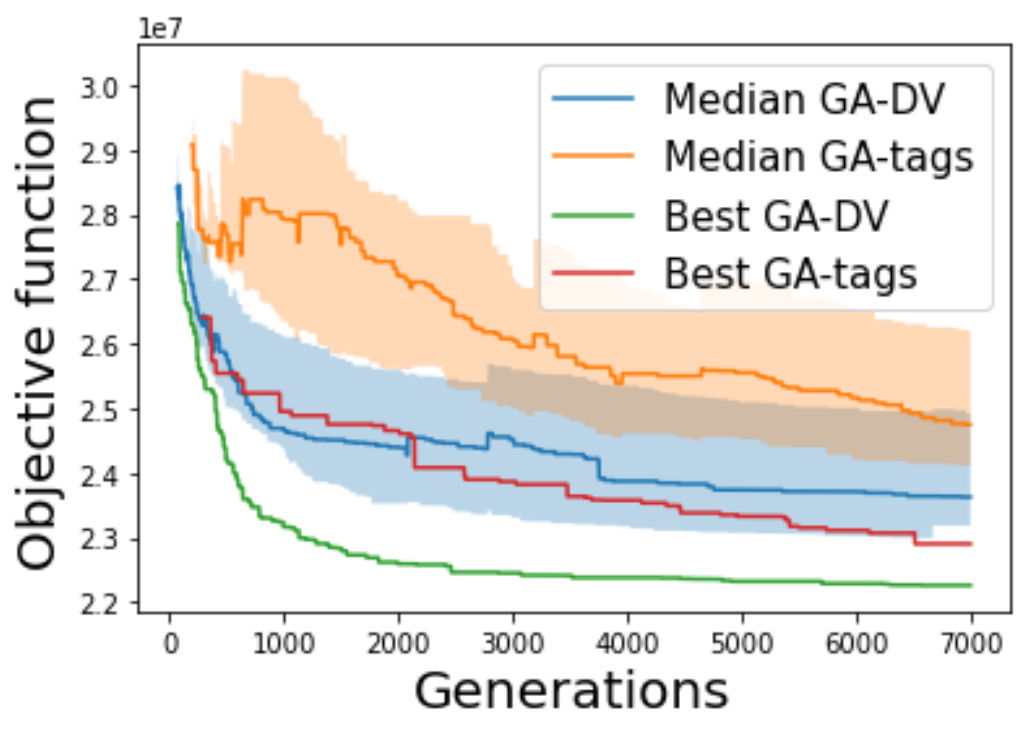} 
  \caption{Occupation rate: 60\%}
  \label{fig:met3}
\end{subfigure}
\begin{subfigure}{1\textwidth}
  \centering
  \includegraphics[width=0.45\linewidth]{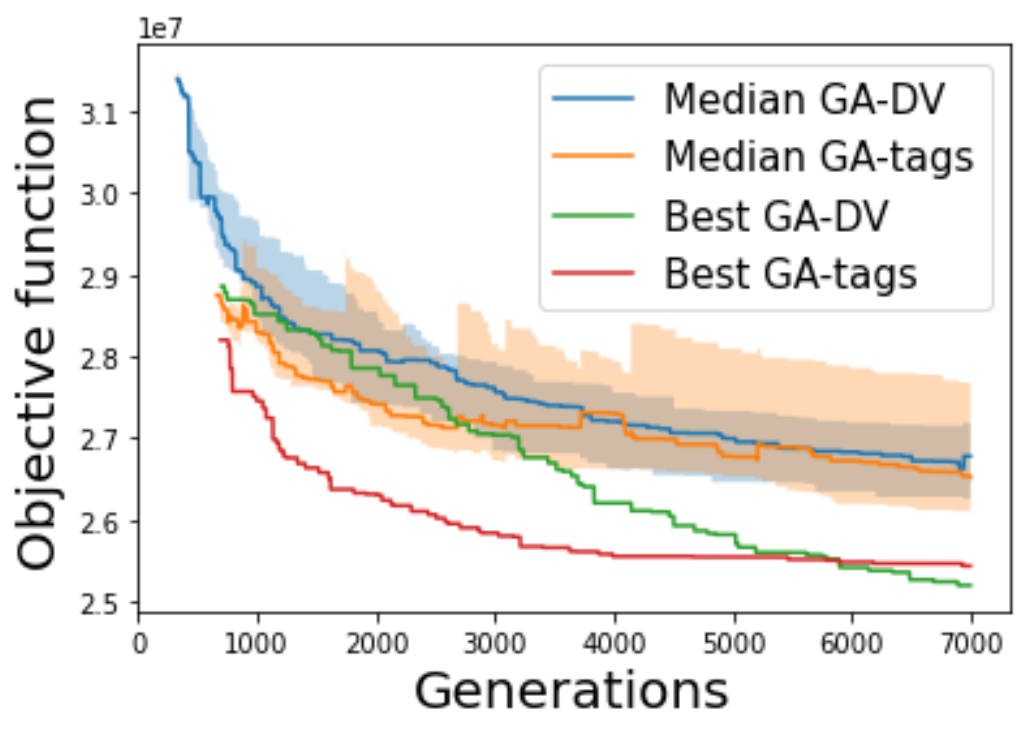} 
  \caption{Occupation rate: 70\%}
  \label{fig:met4}
\end{subfigure}

\caption{Experimental results for both methods (GA-DV and GA-tags) and the five occupation rates.}
\label{fig:metrics}
\end{figure}

Figure \ref{fig:metrics} shows the experimental results for both methods and the five occupation rates of the container. For each occupation rate, the median of the 20 convergence curves obtained for both algorithms are plotted, and the corresponding interquartile is shown. The interquartile corresponds to the gap between the quantiles at 25\% and 75\% to study the dispersion of the results. Furthermore, the best run is plotted as well. For each method, it corresponds to the run (among the 20) which allows to reach the smaller objective function value. 

Table \ref{tab:resultscc} sums up the corresponding numerical results. For both methods and all occupation rates, the number of successful runs are first reported. They correspond to the number of runs that reach a feasible solution. Then, the final median of the objective function as well as the final interquartile range are listed. Finally, the generation corresponding to the apparition of the first feasible solution is specified.

\begin{table}[h!]
\centering
   \begin{tabularx}{1\textwidth} { 
  | >{\centering\arraybackslash}m{.25\textwidth} 
  | >{\centering\arraybackslash}m{.15\textwidth} 
  | >{\centering\arraybackslash}m{.245\textwidth} 
  | >{\centering\arraybackslash}m{.245\textwidth}|}
    \hline
         & OR & GA-tags & GA-DV   \\
        \hline
        \hline
         \multirow{5}{5cm}{Nb of successful runs \\ (a feasible solution is found)}  & 30\% & 20 & 20  \\& 40\% &20 & 20  \\& 50\% &20 & 20  \\ &60\% & 20 & 20 \\ &  70\% & 14 & 15   \\
         \hline
         \multirow{5}{5cm}{Final objective \\ function (median)} & 30\% &  1.55e7 & 1.49e7 \\& 40\% & 1.87e7 & 1.87e7 \\&  50\% & 2.13e7 & 2.078e7   \\ & 60\% &  2.47e7 & 2.36e7  \\&  70\% & 2.65e7 & 2.68e7 \\
         \hline
         \multirow{5}{5cm}{Final interquartile \\ range (IQR) } & 30\% &1.34e6  & 1.91e6   \\& 40\% & 2.33e6 & 2.10e6 \\&  50\% & 9.05e5 & 1.94e6   \\ & 60\% &  1.97e6 & 1.97e6  \\&  70\% & 1.62e6 & 1.92e6  \\
         \hline
         \multirow{5}{5cm}{Generation of \\first feasible solution\\ (mean)} & 30\% & 25.6 & 14.3  \\&  40\% & 92.75 & 40.35 \\&  50\% & 347 & 98.2   \\ & 60\% &  1219.7 & 174.5 \\&  70\% & 2050.2 & 744   \\
         \hline
    \end{tabularx}
    \caption{Numerical results obtained for both GA-tags and GA-DV algorithms.}
    \label{tab:resultscc}
\end{table}

\subsubsection{General analysis}
\label{sec:genan}
First of all, Figure \ref{fig:metrics} and Table \ref{tab:resultscc} highlight that both methods allow to find feasible solutions to the aforementioned VSDS problem. As a matter of fact, an appropriately set up GA enhanced with the proposed mechanism provides a layout of the satellite module among the 3888 possible configurations of components, and manages to solve the constraints for all occupation rates. 

Table \ref{tab:resultscc} shows that until the 60\% occupation rate, 100\% of the runs find a feasible solution from random initializations. For the highest occupation rate, 25\% of them fail to find a feasible solution, regardless of the algorithm. Indeed, the more the occupation rate of the container increases, the more the constraints are difficult to solve. For the same reason, it takes more and more time for both algorithms to find a first feasible solution  as the occupation rate raises, and simultaneously the convergence speed decreases. From 50\%, not all runs seem to have converged by the end of the chosen evaluations cost. 
Finally, the interquartile range (IQR) allows to characterized the robustness of the algorithms. For both algorithms and all the occupation rates, the final IQR is approximately between 5\% and 10\% of the median of the objective function. This indicates a reasonable dispersion of the final results.

Both algorithms seem to behave similarly in terms of convergence. However, some differences may be pointed out regarding the results. Firstly, the GA-DV method reaches a better objective function median for occupation rates below 60\%. At 70\%, the results are more mitigated when it comes to convergence speed and final objective median. 
However, the GA-DV algorithm always provides the best final inertia, regardless of the occupation rate. Moreover, the GA-DV technique allows to find a feasible solution (and so solve the constraints) systematically faster than the GA-tags method. 
At last, the IQR final values are similar and no trend seems to emerge between the two algorithms and over the generations. 

All the results must be contrasted with the fact that they depend on the initializations, even if the algorithms are identically initialized at equal occupation rates. The random initialization allow to keep the general and versatile nature of the methods.

\subsubsection{Constraints handling}
\begin{figure}[ht]
\begin{subfigure}{.32\textwidth}
  \centering
  \includegraphics[width=\linewidth]{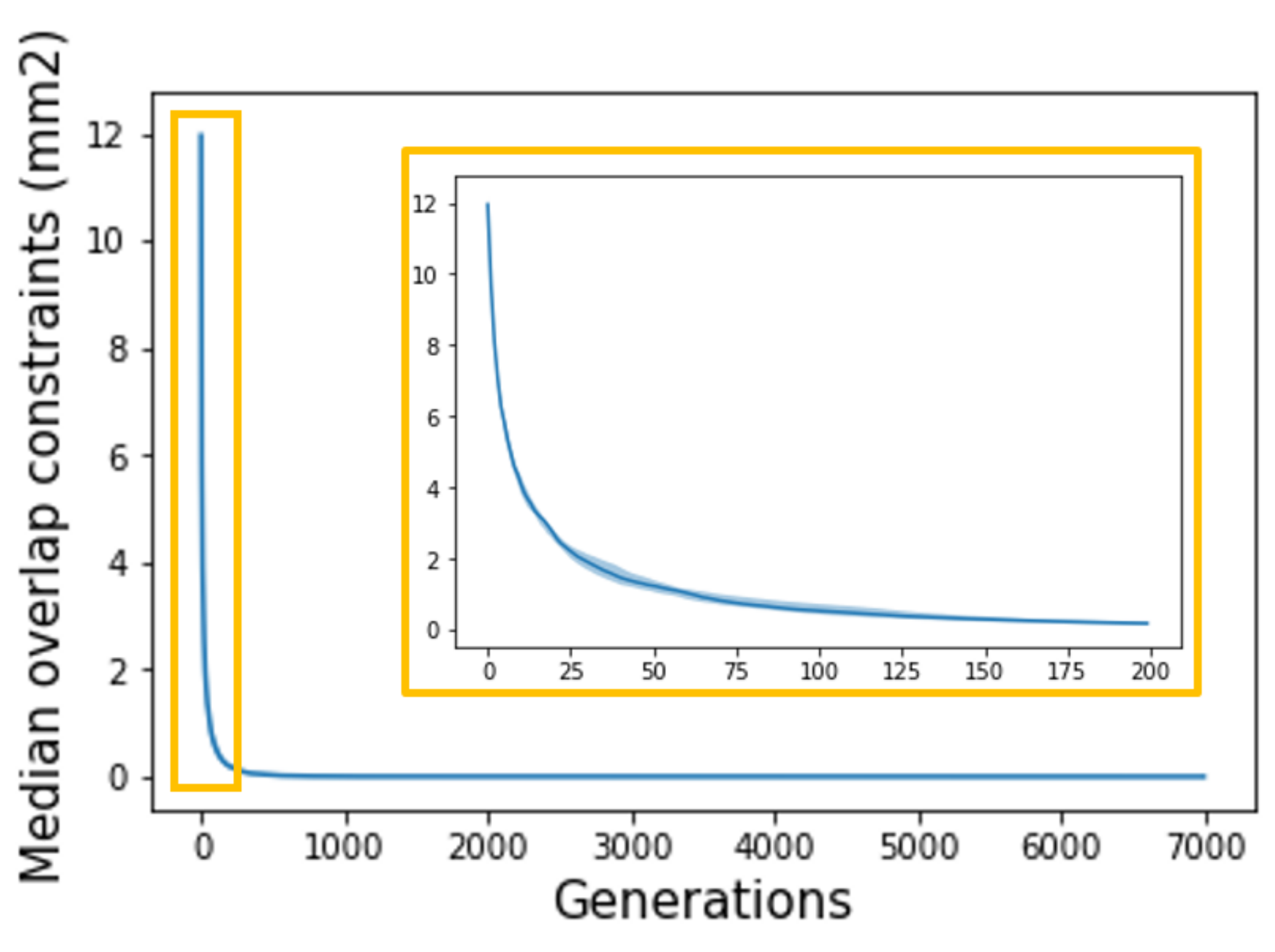}  
  \caption{Overlap constraint (median).}
  \label{fig:cst1}
\end{subfigure}
\begin{subfigure}{.32\textwidth}
  \centering
  \includegraphics[width=\linewidth]{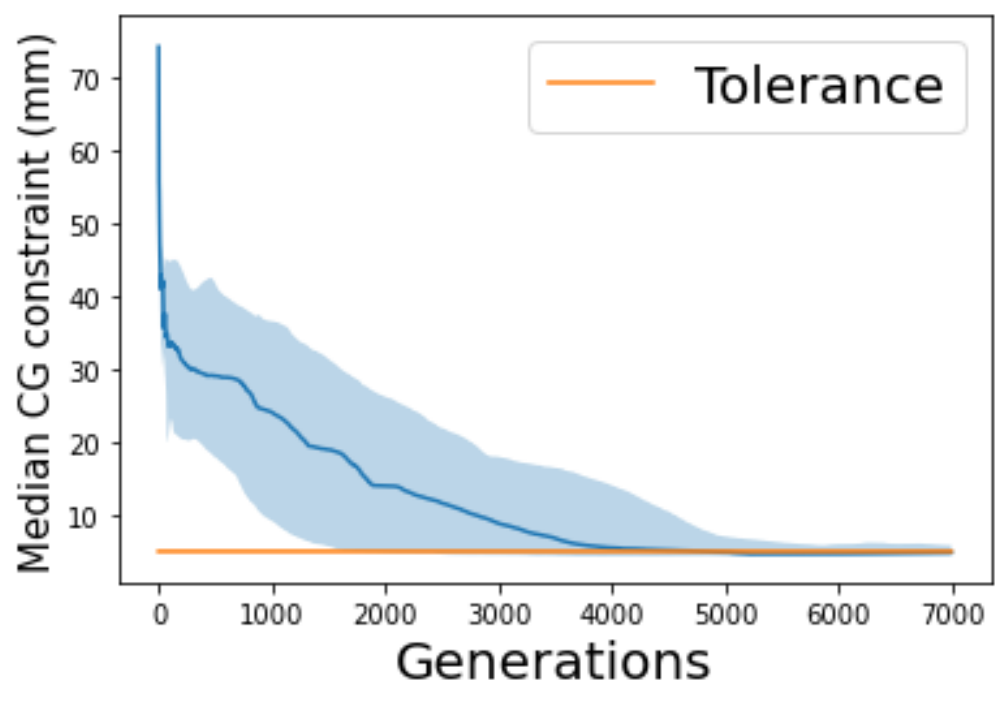} 
  \caption{Centroid constraint (median).}
  \label{fig:cst2}
\end{subfigure}
\begin{subfigure}{.32\textwidth}
  \centering
  \includegraphics[width=\linewidth]{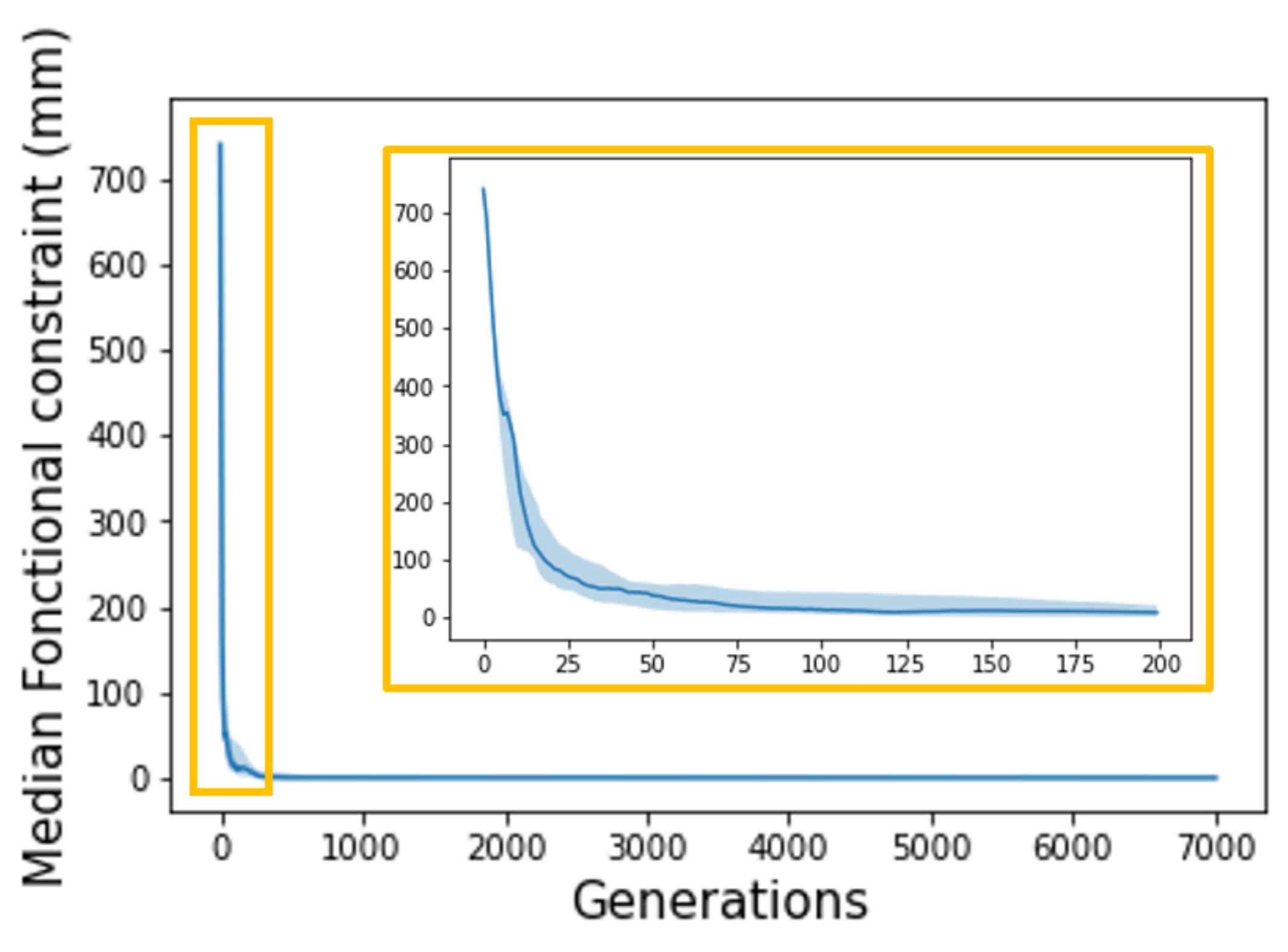} 
  \caption{Functional constraint (median).}
  \label{fig:cst3}
\end{subfigure}
\caption{Evolution of the mean of the constraints in the population for the 70\% compactness and over the 7000 generations.}
\label{fig:cst70}
\end{figure}

In order to understand more deeply the behavior of the algorithms, the resolution of the constraints is analyzed in this section. For the sake of conciseness, the results related to the constraints are only shown on Figure \ref{fig:cst70} for the GA-tags and only for the 70\% occupation rates. This corresponds to the configuration for which the constraints are solved the slowest. 

Figure \ref{fig:cst1} shows the evolution of the median of the overlap constraint among the 20 runs, corresponding to the sum of the overlap constraint between the components and the overlap constraint between components and the exclusion zone. The figure shows the evolution of the overlap constraint over the 7000 generations and show more details over the 200 first generations. 
Figure \ref{fig:cst2} shows the evolution of the median of the center of mass constraint among the 20 runs and for the 7000 generations as well as the tolerance zone. 
Figure \ref{fig:cst3} shows the evolution of the median of the functional constraint among the 20 runs. The figure show the evolution of the overlap constraint over the 7000 generations and show more details over the 200 first generations.

Figure \ref{fig:cst70} demonstrates that the center of mass constraint seems to be the most difficult to solve while the overlap and functional constraints are solved pretty easily even for this high occupation rate. When it comes to the center of gravity constraint, it takes more generations to reach the tolerance zone when the occupation rate increases. For high occupation rates, all the runs corresponding to different initial populations do not find a feasible solution and these runs are non-feasible because of the center of gravity constraint which remains unsolved during the 7000 generations, while the other constraints are respected. This analysis shows that in the general case, the satisfaction of constraints has to be carefully taken into account in order to provide best convergence rates.

\subsubsection{Exploration and diversity}
\begin{figure}[ht]
    \begin{subfigure}{.5\textwidth}
  \centering
  \includegraphics[width=\linewidth]{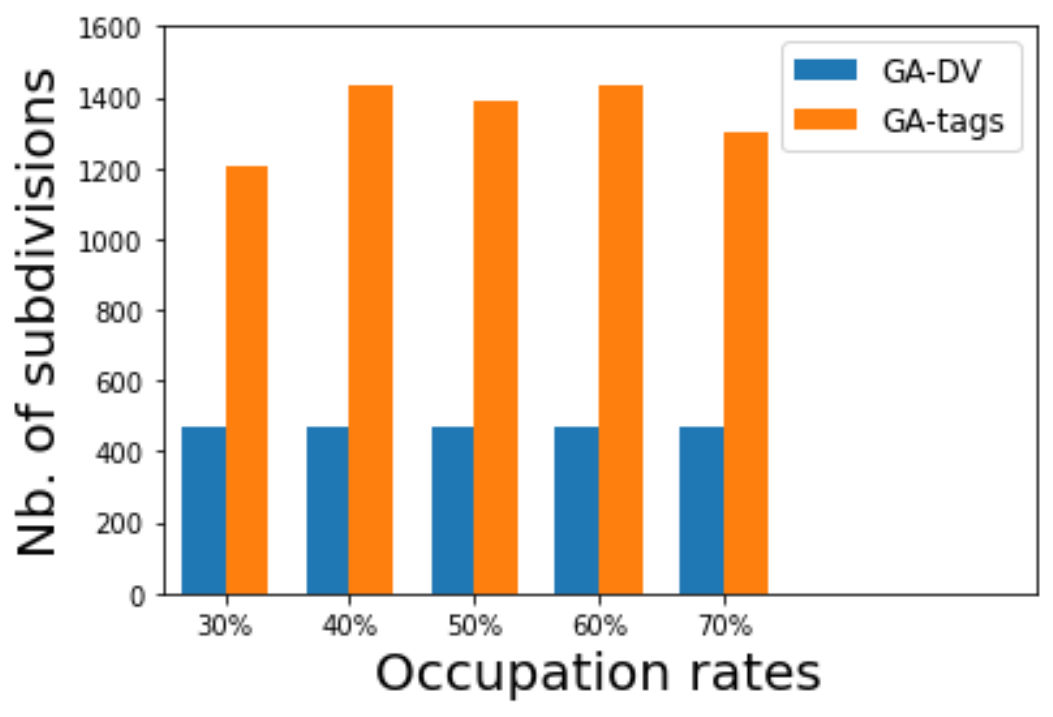}  
  \caption{Among all the individuals.}
  \label{fig:all}
\end{subfigure}
\begin{subfigure}{.5\textwidth}
  \centering
  \includegraphics[width=\linewidth]{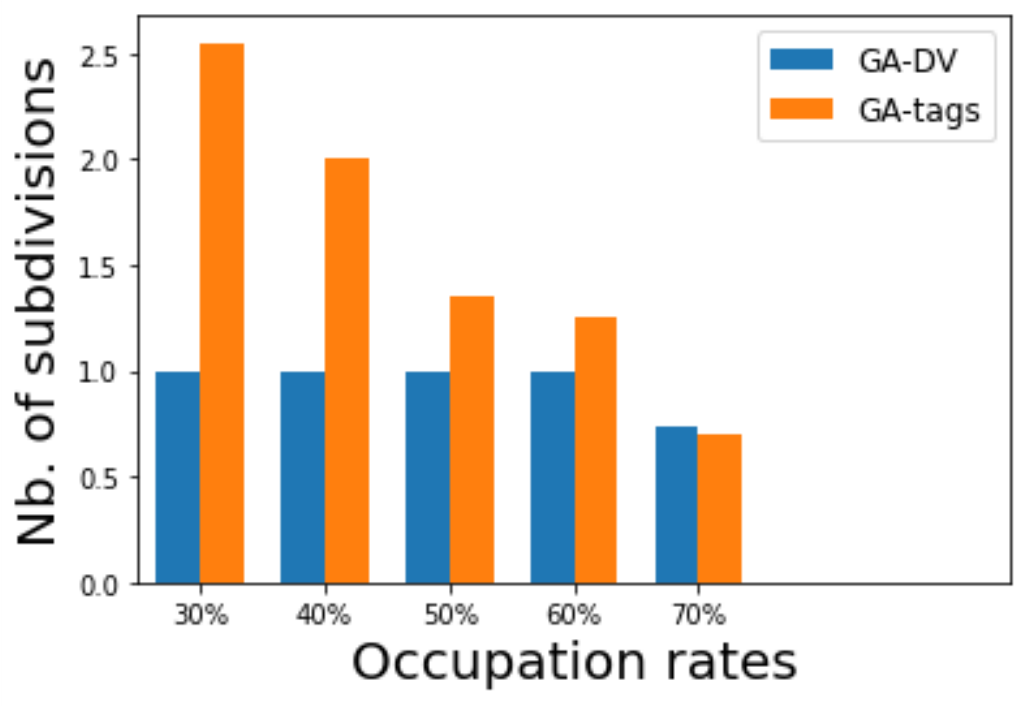} 
  \caption{Among the feasible individuals.}
  \label{fig:feas}
\end{subfigure}
    \caption{Number of configurations reached by both algorithms and for the five occupation rates.}
    \label{fig:configs}
\end{figure}

In order to analyze more in details the exploration behavior of the algorithms, the number of different configurations reached by the methods among the 3888 possible ones will be examined. In more details, for both algorithms and the five occupation rates, the numbers of configurations reached among all individuals and only among the feasible individuals are considered. They are plotted on Figure \ref{fig:configs}.

Figure \ref{fig:all} shows that among all individuals of their populations,  the GA-tags methods explores three times more configurations than the GA-DV method. Moreover, considering only the feasible individuals of their populations, the more the occupation rate increases, the less the GA-tags algorithm seems to explore feasible configurations. For its part, the GA-DV algorithm focuses always on the first feasible configuration found. Those results balance the general conclusions drawn in Section \ref{sec:genan}. Indeed, even if a slightly better convergence tendency was highlighted in favor of the GA-DV method, the better exploration capacities of the GA-tags method must be underlined. It can be explained by the presence of evolutionary operators dedicated to the tag vector responsible for the choices of configurations.

\subsection{Summary of the analyses}

\begin{figure}[ht]
\begin{subfigure}{.32\textwidth}
  \centering
  \includegraphics[width=\linewidth]{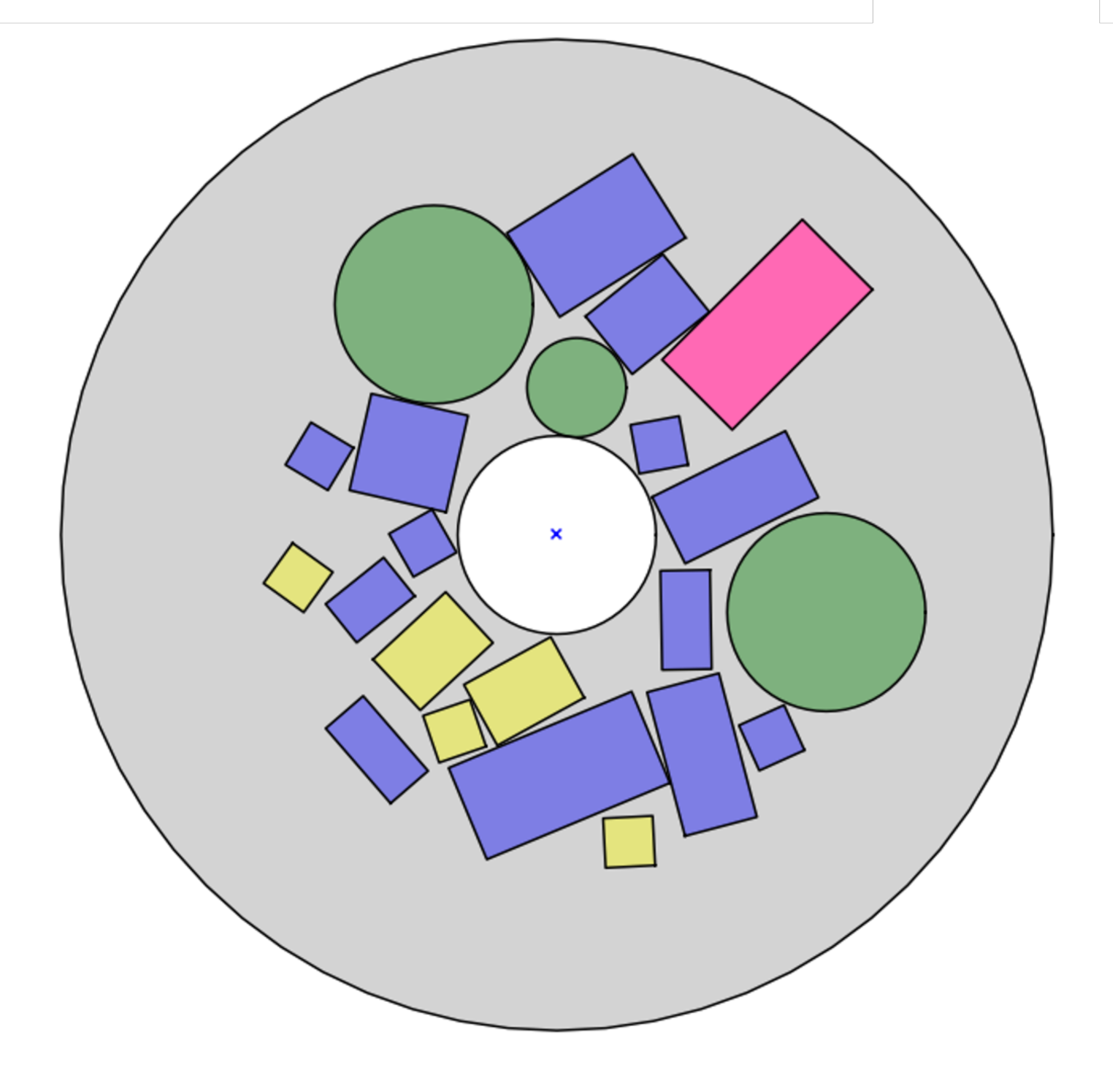}  
  \caption{30\% compactness}
  \label{fig:cv1}
\end{subfigure}
\begin{subfigure}{.32\textwidth}
  \centering
  \includegraphics[width=\linewidth]{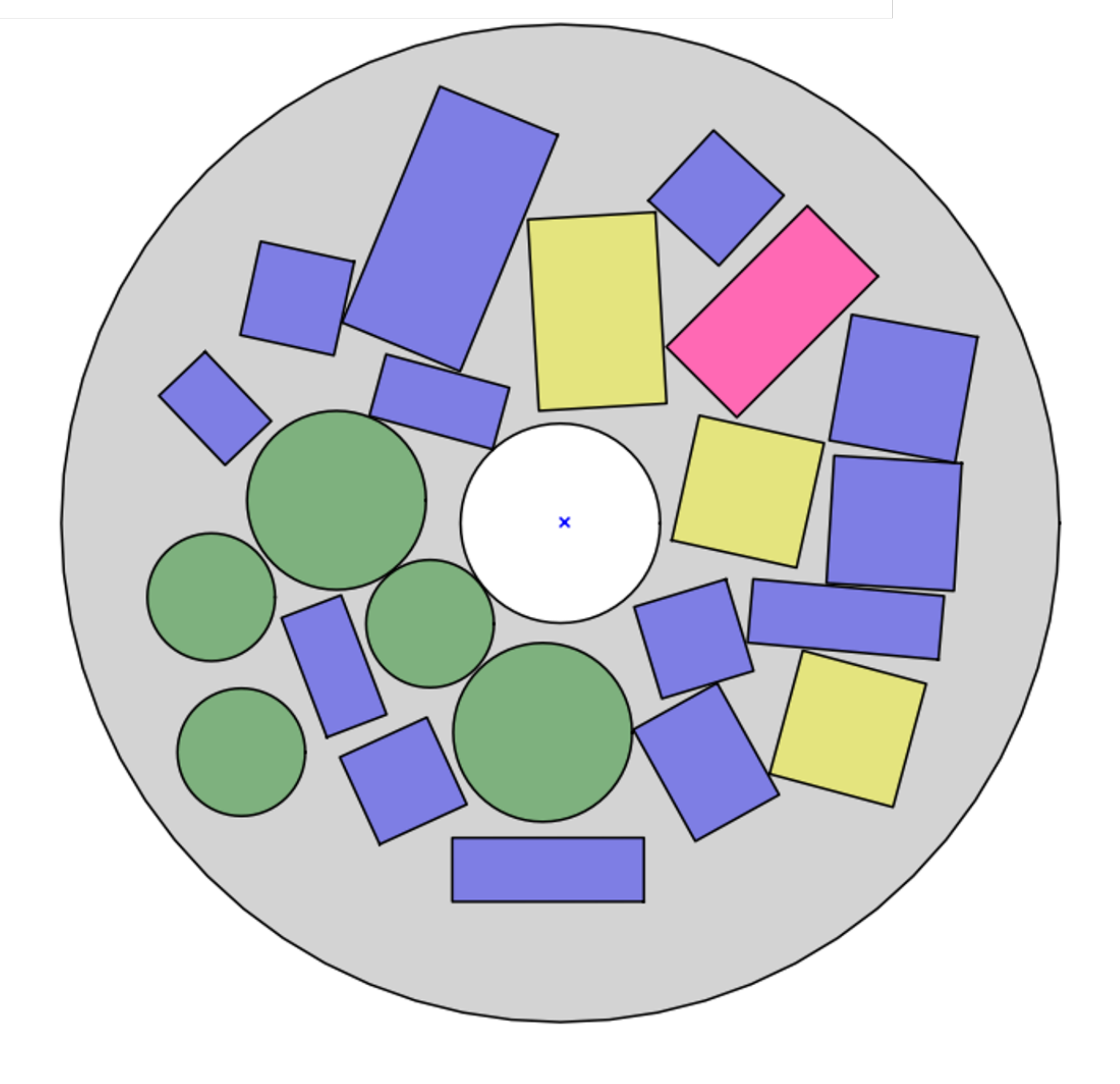} 
  \caption{50\% compactness}
  \label{fig:cv2}
\end{subfigure}
\begin{subfigure}{.32\textwidth}
  \centering
  \includegraphics[width=\linewidth]{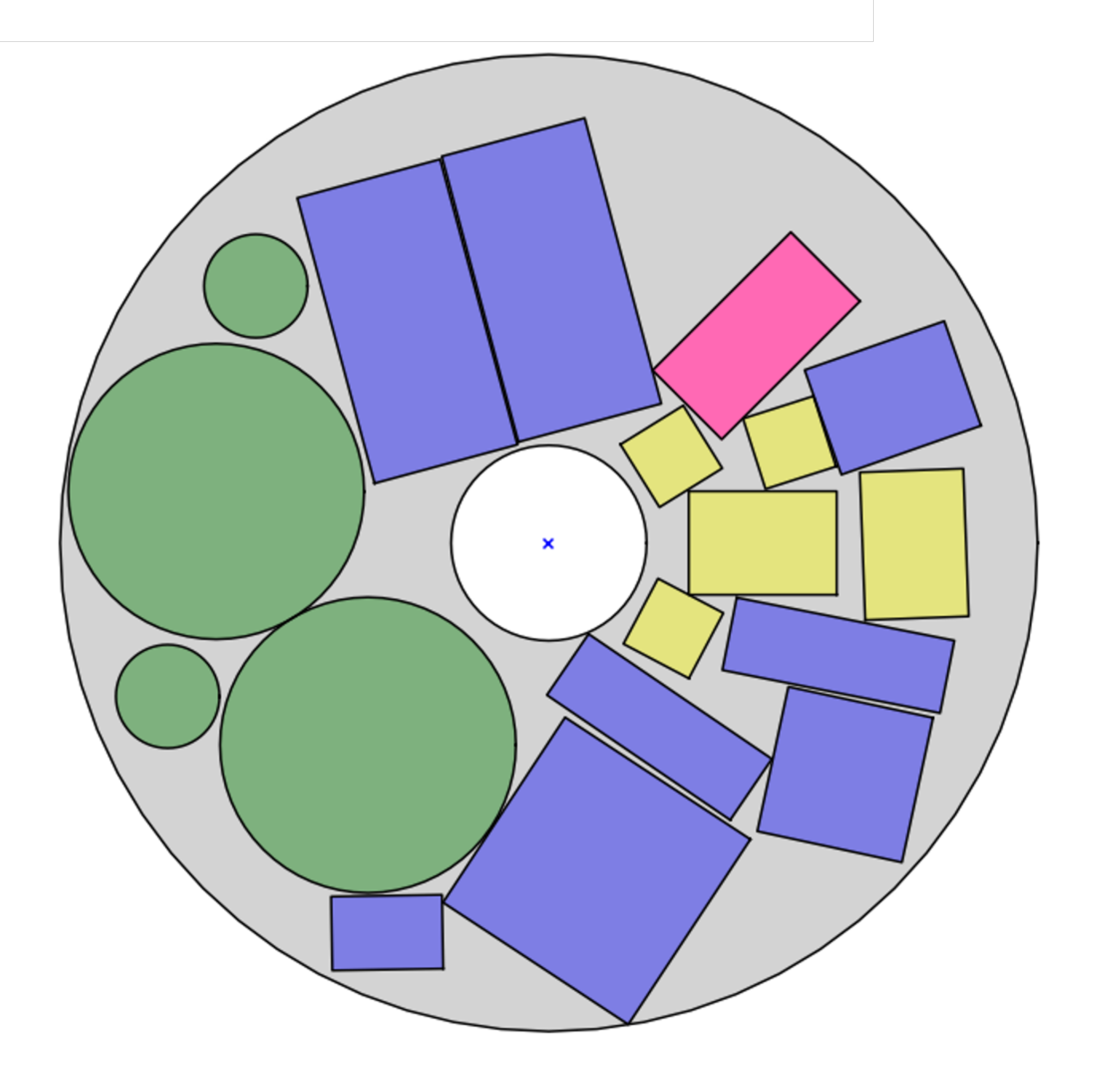} 
  \caption{70\% compactness}
  \label{fig:cv3}
\end{subfigure}
\caption{Optimal layout obtained for the 30\%, 50\% and 70\% occupation rates.}
\label{fig:optimplates}
\end{figure}

Figure \ref{fig:optimplates} shows the best layout found by both algorithms for occupation rates equal to 30\%, 50\% and 70\%.

It is worth noting that for the three different occupation rates different subdivisions of the components have been chosen. In addition, the more the occupation rates increase, the more the algorithms seem to privilege bigger components for the layout.  Globally, both hidden-variable mechanism implementations coupled to a GA manage to find systematically a feasible solution for occupation rates lower than 70\%. However, for both of them, the exploration of feasible configurations is limited and focuses quickly on very few configurations from the time a feasible solution is reached, despite the efforts made to induce diversity trough the operators study. 
Moreover, the more the occupation rate increases, the more the constraints are difficult to solve and consequently less and less runs succeed either in finding a feasible solution during the 7000 generations or in converging enough. This is relatively a classical results for GA \cite{deb2000efficient}. Constraint handling has to be finely tuned to provide best results.

In general, the hidden-variables mechanism allows to extend the range of layout problems to be solved. It enables to tackle a variable-size design space aspect and so to increase the number of architecture choices to be made during the design of the layout phase where a classical GA could only tackle a problem with a fixed number of elements to be placed.

\section{Conclusion}

In this paper, a new formulation of the  optimal layout problem is proposed taking into account a variable-size design space aspect as well as new design variables, new functional constraints, and a new exclusion zone in the design space. This extension enables to describe more realistically engineering problems with the aim to tackle cases of industrial complexity in the future and to extend the range of layout problems to be solved beyond the aerospace vehicles application cases. 
An adapted hidden-variables mechanism has been proposed in order to tackle this new optimal layout problem. It has been applied thanks to two different implementations with a GA (tags and dimensional variables). Both hidden genes mechanisms were compared on an extended realistic satellite optimal layout design problem and a sensitivity analysis has been conducted to the occupation rate of the satellite container to illustrate the performance of the proposed approaches in case of highly constrained problem. The results enable to conclude that both methods revealed convergence capabilities and can be used to solve VSDS problems.  
Further observations lead to focus more in details on the constraints and their resolution, as well as the diversity purpose. 
In the future, more experimentation should be performed on the present algorithms (increase in the occupation rate, variation of the number of configurations of the problem ...). Moreover, the proposed mechanism should be adapted for other evolutionary methods (particle swarm, ant colony), but it is necessary to focus on the adaptations of each evolutionary processes in order to deal properly with the dimensional variables and this was out of the scope of the present paper. Finally, a research for hybridation may be conducted in order to handle easily constraints and diversity.

\section*{Acknowledgments}
\noindent This work is part of a PhD thesis cofunded by INRIA and ONERA.

\newpage

\appendix

\section{Components and decompositions}
\label{sec:appendix0}
\begin{table}[ht]
    \centering
    \begin{tabular}{|c|c|c|c|c|c|}
    \hline
        Components type & Geometry & d1 & d2 & Mass & Subdivisions  \\
        \hline
        \hline
         Fuel & Cylinder & 100 & & 15 & [1,2,3] \\
         \hline
         Fuel & Cylinder& 100 & & 8 & [1,2] \\
         \hline
         Fuel & Cylinder & 50 & & 4 & [1,2] \\
         \hline
         Energy & Cuboid & 200 & 200 & 10 & [1,2,3] \\
         \hline
         Energy & Cuboid & 150 & 100 & 10 & [1,2] \\
         \hline
         Diverse & Cuboid & 100 & 100 & 8 & [1,2,4] \\
         \hline
         Diverse & Cuboid & 150 & 100 & 10 & [1,2] \\
         \hline
         Diverse & Cuboid & 150 & 150 & 12 & [1,2,4] \\
         \hline
         Diverse & Cuboid & 200 & 100 & 15 & [1,2,3] \\
         \hline
         Diverse & Cuboid & 100 & 75 & 6 & [1]\\
         \hline 
         Diverse & Cuboid & 75 & 50 & 2 & [1] \\
         \hline 
         Diverse & Cuboid & 200 & 100 & 15 & [1] \\
         \hline
    \end{tabular}
    \caption{Components and their characteristics for a 30\% compactness}
    \label{tab:decompo}
\end{table}
d1 and d2 are the dimensions of the components (in mm).
The subdivisions are written as [i,j,k] meaning that the corresponding component can be subdivided in three different ways: in i components or j components or k components. The dimensions of those sub-components are obtained such that the area and mass of the initial component are conserved. 

The other compactnesses are obtained by multiplying all the dimensions by the same coefficient to obtain the 50\% and 70\% occupation rates. 

\section{Algorithms configurations for the toy case}
\label{sec:appendix_test}

\begin{table}[ht]
    \centering
    \begin{tabular}{|c|c|c|}
        \hline
         Operators & Choice & Hyper-parameters  \\
         \hline
         \hline
         Constraints Handling & \multicolumn{2}{|c|}{Constraint-Dominance}  \\
         \hline
         Selection & Tournament & 2 individuals \\
         \hline
         Crossover genes & Simulated Binary Crossover & $P_c^{genes} = 0.85$ \\
         \hline
         Crossover tags & 1-point & $P_c^{tags} = 0.85$ \\
         \hline
         Mutation genes & Polynomial Mutation & $P_m^{genes} = 0.25$ \\
         \hline
         Mutation tags & Polynomial Mutation & $P_m^{tags} = 0.4$  \\
         \hline
         Replacement & \multicolumn{2}{|c|}{Non-dominated Truncating}   \\
         \hline
         \end{tabular}
    \caption{GA-tags configuration for the test case.}
    \label{tab:gatagstestt}
\end{table}

\begin{table}[ht]
\centering
   \begin{tabular} { 
  | >{\centering\arraybackslash}m{.25\textwidth} 
  | >{\centering\arraybackslash}m{.25\textwidth} 
  | >{\centering\arraybackslash}m{.25\textwidth} 
  | >{\centering\arraybackslash}m{.25\textwidth} |}
  \hline
         Operators & Choice & GA-DV-num & GA-DV-bin  \\
         \hline
         \hline
         Constraints Handling & \multicolumn{3}{|c|}{Constraint-Dominance}  \\
         \hline
         Selection & Tournament & 2 individuals & 2 individuals \\
         \hline
         Crossover & Simulated Binary Crossover & $P_c^{genes} = 0.85$ & $P_c^{genes} = 0.9$   \\
         \hline
         Mutation & Polynomial Mutation & $P_m^{genes} = 0.35$ & $P_m^{genes} = 0.4$  \\
         \hline
         Replacement & \multicolumn{3}{|c|}{Non-dominated Truncating}   \\
         \hline
  \end{tabular}
  \caption{GA-DV configurations for the test case and the two possible chromosome implementations.}
   \label{tab:gadvtest}
\end{table}

\begin{table}[ht]
    \centering
    \begin{tabular}{|c|c|}
    \hline
        Parameters & value \\
        \hline
        \hline
        Population size & 50 \\
        \hline 
         Number of generations & 60 \\
         \hline
         
    \end{tabular}
    \caption{Numerical configuration for the toy case}
    \label{tab:paramtc}
\end{table}
\newpage

\newpage

 \bibliographystyle{elsarticle-num} 
 \bibliography{manuscript}

\begin{thebibliography}{10}
\expandafter\ifx\csname url\endcsname\relax
  \def\url#1{\texttt{#1}}\fi
\expandafter\ifx\csname urlprefix\endcsname\relax\def\urlprefix{URL }\fi
\expandafter\ifx\csname href\endcsname\relax
  \def\href#1#2{#2} \def\path#1{#1}\fi

\bibitem{deremaux}
Y.~Deremaux, N.~Pietremont, J.~Negrier, E.~Herbin, M.~Ravachol, Environmental
  mdo and uncertainty hybrid approach applied to a supersonic business jet, In
  12th AIAA/ISSMO multidisciplinary analysis and optimization conference,
  Victoria, British Columbia Canada (2008) 5832.

\bibitem{liu}
J.-F. Liu, L.~Hao, G.~Li, Y.~Xue, Z.-X. Liu, J.~Huang, Multi-objective layout
  optimization of a satellite module using the wang-landau sampling method with
  local search, Frontiers of Information Technology \& Electronic Engineering
  17~(6) (2016) 527--542.

\bibitem{cui2019}
F.-Z. Cui, C.-Q. Zhong, X.-K.~W. et~al., A collaborative design method for
  satellite module component assignment and layout optimization, Proceedings of
  the Institution of Mechanical Engineers, Part G: Journal of Aerospace
  Engineering 233(15) (2019) 5471--5491.

\bibitem{wang2009}
Y.-S. Wang, H.-F. Teng, Y.-J. Shi, Cooperative co-evolutionary scatter search
  for satellite module layout design, Engineering Computations 26~(7) (2009)
  761--785.

\bibitem{pelamatti}
J.~Pelamatti, Mixed-variable bayesian optimization, application to aerospace
  system design, PhD thesis Université de Lille (2020).

\bibitem{abdelkhalik2011}
A.~Gad, O.~Abdelkhalik, Hidden genes genetic algorithm for multi-gravity-assist
  trajectories optimization, AIAA Journal of Spacecraft and Rockets 48(4)
  (2011) 629--641.

\bibitem{cagan}
J.~Cagan., S.~Szykman, Constrained three-dimensional component layout using
  simulated annealing., ASME. Journal of Mechanical Design 119 (1) (1997)
  28--35.

\bibitem{singh}
S.~P. Singh, R.~R.~K. Sharma, A review of different approaches to the facility
  layout problems, The International Journal of Advanced Manufacturing
  Technology 30(5) (2006) 425--433.

\bibitem{meller}
R.~D. Meller, Y.~A. Bozer, A new simulated annealing algorithm for the facility
  layout problem, International Journal of Production Research 34:6 (1996)
  1675--1692.

\bibitem{jacquenot}
T.~G. Jacquenot, {Méthode générique pour l’optimisation d’agencement
  géométrique et fonctionnel}., PhD thesis (in French), Ecole Centrale de
  Nantes (ECN) (2010).

\bibitem{liu2018}
J.~Liu, H.~Zhang, K.~He, S.~Jiang, Multi-objective particle swarm optimization
  algorithm based on objective space division for the unequal-area facility
  layout problem, Expert Systems With Applications 102 (2018) 179--192.

\bibitem{derakhshan}
A.~D. Asl, K.~Y. Wong, M.~K. Tiwari, Unequal-area stochastic facility layout
  problems: solutions using improved covariance matrix adaptation evolution
  strategy, particle swarm optimisation, and genetic algorithm, International
  Journal of Production Research 54(3) (2016) 799--823.

\bibitem{hasan}
R.~A. Hasan, M.~A. Mohammed, N.~Ţăpuş, O.~A. Hammood, A comprehensive study:
  Ant colony optimization (aco) for facility layout problem, 16th RoEduNet
  conference: networking in education and research (RoEduNet) IEEE (2017) 1--8.

\bibitem{potter94}
Potter, A.~Mitchell, D.~Jong, A.~Kenneth, A cooperative coevolutionary approach
  to function optimization, International Conference on Parallel Problem
  Solving from Nature Springer, Berlin, Heidelberg (1994) 249--257.

\bibitem{ma2018}
X.~Ma, X.~LI, Q.~Z. et~al., A survey on cooperative co-evolutionary algorithms,
  IEEE Transactions on Evolutionary Computation 23 (2018) 421--441.

\bibitem{hong2010}
H.~fei Teng, Y.~Chen, W.~Z. et~al., A dual-system variable-grain cooperative
  coevolutionary algorithm: satellite-module layout design, IEEE transactions
  on evolutionary computation 14 (3) (2009) 438--455.

\bibitem{lim2016}
Z.~Y. Lim, S.~Ponnambalam, et~Kazuhiro~Izui, Multi-objective hybrid algorithms
  for layout optimization in multi-robot cellular manufacturing systems,
  Knowledge-Based Systems 120 (2017) 87--98.

\bibitem{li2016}
Z.~Li, Y.~Zeng, Y.~Wang, L.~Wang, B.~Song, A hybrid multi-mechanism
  optimization approach for the payload packing design of a satellite module,
  Applied Soft Computing 45 (2016) 11--26.

\bibitem{liu2012}
J.~Liu, Constrained layout optimization in satellite cabin using a multiagent
  genetic algorithm, Asia-Pacific Conference on simulated evolution and
  learning, Springer, Berlin, Heidelberg, Deutschland (2012) 440--449.

\bibitem{xu}
Y.~C. Xu, R.~B. Xiao, M.~Amos, A novel genetic algorithm for the layout
  optimization problem, 2007 IEEE Congress on Evolutionary Computation, Tokyo,
  Japan (2007) 3938--3943.

\bibitem{xu2010}
Y.~C. Xu, F.~M. Dong, Y.~Liu, R.~B. Xiao, M.~Amos, Ant colony algorithm for the
  weighted item layout optimization problem, arXiv preprint:1001.4099 (2010).

\bibitem{oliveira}
G.~F.~O. Alves, J.~C. Becceneri, S.~Sandri, A balancing heuristic for
  spacecraft equipment layout optimization, 2015 IEEE International Conference
  on Fuzzy Systems (FUZZ-IEEE), Istanbul, Turkey (2015) 1--8.

\bibitem{tarkesh2009}
H.~Tarkesh, A.~Atighehchian, A.~S. Nookabadi, Facility layout design using
  virtual multi-agent system, Journal of Intelligent Manufacturing 20 (4)
  (2009) 347.

\bibitem{burggraf2021}
P.~Burggraäf, J.~Wagner, B.~Heinbach, Bibliometric study on the use of machine
  learning as resolution technique for facility layout problems, IEEE Access 9
  (2021) 22569--22586.

\bibitem{tsuchiya1996}
K.~Tsuchiya, S.~Bharitkar, Y.~Takefuji, A neural network approach to facility
  layout problems, European Journal of Operational Research 89 (3) (1996)
  556--563.

\bibitem{vashisht}
D.~Vashisht, H.~Rampal, H.~Liao, Y.~Lu, D.~Shandbhag, E.~Fallon, L.~B. Kara,
  Placement inintegrated circuits using cyclic reinforcement learning and
  simulated annealing, arXiv preprintarXiv:2011.07577 (2020).

\bibitem{mirhoseini}
A.~Mirhoseini, A.~Goldie, M.~Yazgan, J.~W. Jiang, E.~Songhori, S.~Wang,
  J.~Dean, A graph placement methodology for fast chip design, Nature 594(7862)
  (2021) 207--212.

\bibitem{cheng}
R.~Cheng, J.~Yan, On joint learning for solving placement and routing in chip
  design, arXiv preprint arXiv:2111.00234. (2021).

\bibitem{dasgupta}
D.~Dasgupta, D.~R. McGregor, Non-stationary function optimization using the
  structured genetic algorithm, Parallel Problem Solving from Nature (PPSN-2)
  Conference, Brussels, Belgium 2 (1992) 145--154.

\bibitem{gentile}
L.~Gentile, C.~Greco, E.~Minisci, T.~Bartz-Beielstein, Structured-chromosome ga
  optimisation for satellite tracking, Proceedings of the Genetic and
  Evolutionary Computation Conference Companion, Prague, Czech Re- public
  (2019) 1955--1963.

\bibitem{hutt2007}
B.~Hutt, K.~Warwick, Synapsing variable-length crossover: Meaningful crossover
  for variable-length genomes, IEEE transactions on evolutionary computation 11
  (1) (2007) 118--131.

\bibitem{abdelpop}
O.~Abdelkhalik, A.~Gad, Dynamic-size multiple populations genetic algorithm for
  multigravity-assist trajectory optimization, Journal of Guidance, Control,
  and Dynamics 35(2) (2012) 520--529.

\bibitem{xue}
Y.~Xue, B.~Xue, M.~Zhang, Self-adaptive particle swarm optimization for
  large-scale feature selection in classification, ACM Transactions on
  Knowledge Discovery from Data (TKDD) 13(5) (2019) 1--27.

\bibitem{xue2}
Y.~Xue, T.~Tang, W.~Pang, A.~X. Liu, Self-adaptive parameter and strategy based
  particle swarm optimization for large-scale feature selection problems with
  multiple classifiers, Applied Soft Computing 88 (2020) 106031.

\bibitem{ellithy}
A.~Ellithy, O.~Abdelkhalik, J.~Englander, Multi-objective hidden genes genetic
  algorithm for multigravity-assist trajectory optimization, Journal of
  Guidance, Control, and Dynamics (2022) 1--17.

\bibitem{liu2008}
Z.~W. Liu, H.~F. Teng, Human–computer cooperative layout design method and
  its application, Computers \& Industrial Engineering 55(4) (2008) 735--757.

\bibitem{sun}
Z.~G. Sun, H.~F. Teng, Optimal layout design of a satellite module, Engineering
  Optimization 35(5) (2003) 513--529.

\bibitem{deb2000}
K.~Deb, An efficient constraint handling method for genetic algorithms,
  Computer methods in applied mechanics and engineering 186(2-4) (2000)
  311--338.

\bibitem{deb}
K.~Deb, R.~B. Agrawal, Simulated binary crossover for continuous search space,
  Complex systems 9 (2) (1995) 115--148.

\bibitem{talbi}
E.-G. Talbi, Metaheuristics: from design to implementation, John Wiley \& Sons
  (2009).

\bibitem{taguchi}
R.~K. Roy, A primer on the taguchi method, Society of Manufacturing Engineers
  (2010).

\bibitem{runarsson}
T.~P. Runarsson, X.~Yao, Stochastic ranking for constrained evolutionary
  optimization, IEEE Transactions on Evolutionary Computation 4 (3) (2000)
  284--294.

\bibitem{deb2002}
K.~Deb, A.~Pratap, S.~Agarwal, T.~Meyarivan, A fast and elitist multiobjective
  genetic algorithm: Nsga-ii, IEEE transactions on evolutionary computation
  6(2) (2002) 182--197.

\bibitem{deb2000efficient}
K.~Deb, An efficient constraint handling method for genetic algorithms,
  Computer methods in applied mechanics and engineering 186~(2-4) (2000)
  311--338.

\end{thebibliography}





\end{document}